\documentclass{article}

\usepackage[nonatbib,final]{neurips_2024}

\usepackage{graphicx}
\usepackage{url} 
\usepackage{multirow}
\usepackage{xcolor}
\usepackage{amsmath}
\usepackage{subcaption}
\usepackage{wrapfig}
\usepackage{array}
\usepackage{ragged2e}
\usepackage{colortbl}
\usepackage{enumitem}
\newcolumntype{L}[1]{>{\raggedright\let\newline\\\arraybackslash\hspace{0pt}}m{#1}}
\newcolumntype{C}[1]{>{\centering\let\newline\\\arraybackslash\hspace{0pt}}m{#1}}
\newcolumntype{R}[1]{>{\raggedleft\let\newline\\\arraybackslash\hspace{0pt}}m{#1}}

\usepackage[pagebackref,breaklinks,colorlinks,bookmarks=false,urlcolor=black,citecolor={green!60!black}]{hyperref}

\usepackage{utfsym}
\usepackage{amsthm}
\usepackage{amsfonts}

\newtheorem{theorem}{Theorem}
\newtheorem{corollary}{Corollary}
\newtheorem{lemma}{Lemma}

\newtheorem{definition}{Definition}

\usepackage{booktabs}

\usepackage{xspace}
\makeatletter
\DeclareRobustCommand\onedot{\futurelet\@let@token\@onedot}
\def\@onedot{\ifx\@let@token.\else.\null\fi\xspace}
\def\eg{\emph{e.g}\onedot} 
\def\ie{\emph{i.e}\onedot} 
 
\def\etc{\emph{etc}\onedot} 
 
\def\etal{\emph{et al}\onedot}
\makeatother

\newcommand{\heart}{$\;\!$\usym{2665}}

\title{Graph Self-Supervised Learning with Learnable Structural and Positional Encodings}

\author{%
Asiri Wijesinghe$^{\dagger}$ \quad Hao Zhu$^{\dagger}$ \quad Piotr Koniusz\thanks{The corresponding author.$\quad$The code is available at: 
\url{https://github.com/wokas36/StructPosGSSL}
}$\;^{,\dagger,\S}$\\
$^{\dagger}$Data61$\!${\color{red}\heart}CSIRO \quad $^\S$Australian National University\\
{\tt\small \{asiriwijesinghe.wijesinghe, allen.zhu, piotr.koniusz\}@data61.csiro.au}
}

\begin{document}

\maketitle

\begin{abstract}
Traditional Graph Self-Supervised Learning (GSSL) struggles to capture complex structural properties well. This limitation stems from two main factors: (1) the inadequacy of conventional Graph Neural Networks (GNNs) in representing sophisticated topological features, and (2) the focus of self-supervised learning solely on final graph representations. To address these issues, we introduce \emph{GenHopNet}, a GNN framework that integrates a $k$-hop message-passing scheme, enhancing its ability to capture local structural information without explicit substructure extraction. We theoretically demonstrate that \emph{GenHopNet} surpasses the expressiveness of the classical Weisfeiler-Lehman (WL) test for graph isomorphism. Furthermore, we propose a structural- and positional-aware GSSL framework that incorporates topological information throughout the learning process. This approach enables the learning of representations that are both sensitive to graph topology and invariant to specific structural and feature augmentations. Comprehensive experiments on graph classification datasets, including those designed to test structural sensitivity, show that our method consistently outperforms the existing approaches and maintains computational efficiency. Our work significantly advances GSSL's capability in distinguishing graphs with similar local structures but different global topologies.
\end{abstract}

\section{Introduction}
Graph Neural Networks (GNNs) are powerful deep learning networks for graph-structured data, employed by various tasks \cite{kipf2016semi, velivckovic2018graph, you2020l2, xu2018powerful, uai_ke, zhang2020deep, wu2020comprehensive, Koniusz2020PowerNI,Prabowo2023MessagePN, ju2024comprehensive,ANAND2024134456,li2025inductive,ding2025lego}. While most GNNs focus on semi-supervised learning, Self-Supervised Learning (SSL)  learns graph representations without human annotations.

Graph Self-Supervised Learning (GSSL)  often outperforms supervised methods in both node-level and graph-level downstream tasks \cite{xu2021infogcl, zhu2021graph, zhu2021refine,zhang2022costa,zhu2021contrastive,glen,zhang2023spectral,zhang2024geometric,hyper_collapse}. In this paper, we focus on graph classification, a crucial graph-level task with significant applications in areas such as molecular property prediction, social network analysis, and protein function classification \cite{ hamilton2017representation, xu2018powerful, feng2022powerful, zhang2023mitigating, fitness_hao}. Graph classification presents unique challenges compared to node-level tasks as it must capture global structural information across different graphs, not just local neighborhoods. Graphs can vary significantly in size and structure, demanding more flexible and expressive models. To obtain effective graph-level representations, models must aggregate information from all nodes and edges while preserving discriminative structural features.

Despite GSSL’s success, they often fail to fully leverage the expressive power of GNNs, by not utilizing both topological and positional information for graph classification. Topological information captures the local structural relationships within the graph through the $k$-hop neighborhood substructure patterns (\eg, triangles, cycles), while positional information, derived from Laplacian eigenvectors or random-walk diffusion, reflects the nodes' relative positions within the graph's global structure.  The lack of topological and positional focus prevents GSSL from distinguishing between graphs with similar local structures but different global topologies. Specifically, in graphs where nodes may have identical local structures (\eg, isomorphic or symmetrical nodes), relying only on neighboring features is inadequate for differentiation. Positional information is critical for enabling GNNs to distinguish such nodes, even when their connectivity patterns are similar. Isomorphic nodes, which cannot be differentiated based solely on their structural information, present a particular challenge. By incorporating positional encodings, GNNs can leverage this additional context to break symmetry, facilitating better differentiation among nodes. This enhancement improves the model's recognition of unique identities, leading to more accurate predictions in graph-related tasks.

The limitations of current GSSL methods can be attributed to two main factors: GNN Architecture Limitations and Self-Supervised Learning Constraints. Conventional GNNs typically aggregate information from immediate neighborhoods, often missing crucial structural differences that exist beyond local structures. For instance, GIN \cite{xu2018powerful} has shown that certain GNN-based methods \cite{kipf2016semi, velivckovic2018graph} are less effective at distinguishing graph structures compared to Weisfeiler-Lehman (WL) based methods. Furthermore, current GSSL methods \cite{velickovic2019deep, sun2019infograph, hassani2020contrastive, you2020graph} often fail to fully leverage the complementary nature of structural and positional information, which hinders their ability to differentiate non-isomorphic graphs with similar local attributes but different global topologies.

Building on these insights, we develop a framework that fundamentally reimagines graph representation learning by innovating both  GNN architecture and the self-supervised learning process. Our goal is to significantly enhance the expressiveness and representational capacity of GSSL in distinguishing non-isomorphic graphs with similar local structures but different global topologies. 
To this end, we focus on two main components: 

%
\renewcommand{\labelenumi}{\arabic{enumi}.}
\begin{enumerate}[leftmargin=0.6cm]
    \item \emph{GenHopNet GNN.} A novel GNN architecture designed to capture complex structural information beyond immediate neighborhoods by a $k$-hop message-passing scheme that expands the receptive field of each node, letting the model  capture long-range dependencies and global structural information. 

\item {\em Structural- and Positional-aware Self-Supervised Learning.} A new self-supervised learning framework that preserves and uses crucial topological information by incorporating both structural and positional information into  learning to overcome the sole focus on final graph representations. 
\end{enumerate}

\vspace{0.1cm}
\noindent
\textbf{Contributions.} Below we summarize our main contributions:
\renewcommand{\labelenumi}{\roman{enumi}.}
\begin{enumerate}[leftmargin=0.6cm]
    \item We introduce \emph{GenHopNet}, a GNN framework that implements a $k$-hop message-passing aggregation scheme and surpasses the expressiveness of the WL test.
    \item We propose a structural- and positional-aware GSSL framework, \emph{StructPosGSSL}, for GNN pre-training, enabling the learning of representations invariant to specific structural and feature augmentations while preserving topological and positional information.
    \item With extensive experiments on both real-world and synthetic datasets we demonstrate that our \emph{StructPosGSSL} achieves superior performance on most graph classification benchmarks.
\end{enumerate}

\begin{figure*}
    \centering
    	\begin{subfigure}[t]{0.73\linewidth} 
		\centering
    \includegraphics[width=1\textwidth]{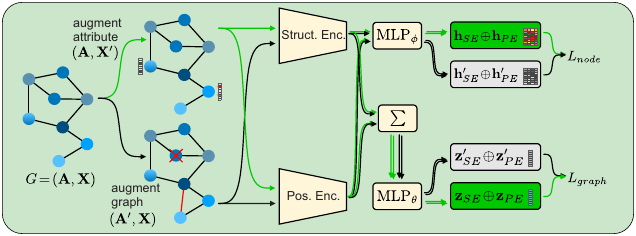}
    \vspace{-0.4cm}
		\vspace{0.1cm}
		\caption{}
		\label{fig_1_1}
	\end{subfigure}
    \hspace{0.2cm}
     \begin{subfigure}[t]{0.22\linewidth} 
		\centering
    \includegraphics[width=0.95\textwidth]{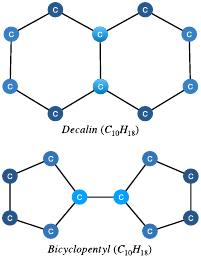}
		\vspace{0.1cm}
		\caption{}
		\label{fig_1_2}
	\end{subfigure}
    \vspace{-0.1cm}
    \caption{({\em a}) A high-level overview of the architecture of \emph{StructPosGSSL} ($G$ is an input graph with two views $(\mathbf{A},\mathbf{X}')$ and $(\mathbf{A}',\mathbf{X})$). Our design comprises three main components: (i) a structural encoder ({\em Struct. Enc.}) that generates  structural embeddings ($\mathbf{h}_{SE}$ and $\mathbf{h}_{SE}'$) for nodes based on their local structural properties; (ii) a positional encoder ({\em Pos. Enc.}) that generates positional embeddings ($\mathbf{h}_{PE}$ and $\mathbf{h}_{PE}'$) for nodes; and (iii) a node aggregation layer, $\sum$, acting over all node embeddings to generate a global graph representation. Moreover, $\text{MLP}_{\phi}$ and $\text{MLP}_{\theta}$ are two projection heads for node representations and graph representations, whereas $\oplus$ is concatenation. ({\em b}) The real-world graph structures of two molecules, Decalin and Bicyclopentyl. While standard Graph SSL frameworks cannot distinguish between these molecular structures, our model can  differentiate them.}
    \label{fig:model-architecture}
\end{figure*}

\section{Related Work}\label{sec:background}
GNNs are designed to effectively process and represent graph-structured data, and they come in various flavors, including GCN \cite{kipf2016semi}, GAT \cite{velivckovic2018graph}, GraphSAGE \cite{hamilton2017inductive}, GIN \cite{xu2018powerful}, DFNets \cite{wijesinghe2019dfnets}, linear SSGC \cite{zhu2021simple}, GReLU \cite{zhang2022graph}, \etc. Such models distinguish representations of graphs based on their data labels. However, annotating graph data, such as identifying categories of biochemical molecules, often requires specialized expertise, making it challenging to obtain large-scale labeled graph datasets \cite{you2020graph}. This challenge highlights a key limitation of supervised graph representation learning.

Contrastive Learning (CL) stands out as a highly effective self-supervised technique embedding unlabeled data \cite{li2022metamask}. By bringing similar examples closer together and pushing dissimilar ones apart, CL methods including SimCLR \cite{chen2020simple}, MoCo \cite{he2020momentum}, BYOL \cite{grill2020bootstrap}, MetAug \cite{li2022metaug}, Barlow Twins \cite{zbontar2021barlow} and their multi-head variants \cite{multihead_lei} have demonstrated remarkable success in  computer vision \cite{wu2018unsupervised, qiang2022interventional}.

\vspace{0.1cm}
\noindent
\textbf{Graph Self-Supervised Learning (GSSL)} 
is a promising technique for learning representations of graph-structured data without requiring labeled examples, making it especially effective for graph classification tasks. To date, many GSSLs with unique strategies have been proposed to enhance  graph classification. These methods build on the strengths of GNNs and CL techniques \cite{velickovic2019deep, suninfograph, hassani2020contrastive}.

A key focus of GSSL is the development of effective graph augmentation strategies. For instance, GraphCL \cite{you2020graph} introduces perturbation invariance and proposes various graph augmentations, such as node dropping, edge perturbation, attribute masking, and subgraph extraction. Recognizing the limitations of using complete graphs, Subg-Con \cite{jiao2020sub} advocates for subgraph sampling as a more effective method for capturing structural information. To improve the semantic depth of sampled subgraphs, MICRO-Graph \cite{zhang2020motif} proposes generating informative subgraphs by learning graph motifs. Furthermore, the process of selecting suitable graph augmentations can be time-consuming and labor-intensive; JOAO \cite{you2021graph} addresses this by introducing a bi-level optimization framework that automates the selection of data augmentations tailored to specific graph data. RGCL \cite{li2022let} argues that random destruction of graph properties during augmentation can lead to a loss of critical semantic information and proposes a rationale-aware approach for graph augmentation. Additionally, SPAN \cite{linspectral} introduces a spectral perspective for guiding topology augmentation,
while SFA \cite{zhang2023spectral} is a spectral embedding augmentation. 
%
%
To capture the hierarchical structures by GSSL, HGCL \cite{ju2023unsupervised} proposes Hierarchical GSSL, which integrates node-level CL, graph-level CL, and mutual CL components. HDC \cite{hyper_collapse} studies hierarchical dimensional collapse in hyperbolic spaces. Another important aspect of GSSL is the process of negative sampling; BGRL \cite{thakoorlarge} simplifies this process by eliminating the need for constructing negative samples, allowing it to scale efficiently to large graphs. To mitigate sampling bias, PGCL \cite{lin2022prototypical} introduces a negative sampling strategy based on semantic clustering. 
COLES \cite{zhu2021contrastive} reformulate Laplacian eigenmaps into CL while GLEN \cite{glen} formulates COLES as the rank difference optimization. COSTA \cite{zhang2022costa} uses sketching to create embedding perturbations.
In contrast, we emphasize both global and local structural understanding. Global graph representations  capture complex topological similarities and differences, while local node embeddings are refined to preserve detailed structural and positional nuances. By incorporating structural and positional awareness through invariance, variance, and covariance across node features, our method improves the ability to distinguish between isomorphic and non-isomorphic graphs. 

\vspace{0.1cm}
\noindent
\textbf{Enhancing GNN Expressiveness. } A substantial amount of effort has been devoted to enhancing the expressive power of GNNs beyond the 1-WL\footnote{WL stands for the Weisfeiler Leman graph isomorphism test.}. This pursuit arises from the need to capture more intricate graph structures and relationships to address complex real-world problems effectively. Broadly, there are four primary directions which GNNs can extend beyond the 1-WL level: (1) A number of studies have introduced higher-order variants of GNNs, demonstrating comparable expressiveness to k-WL with $k\geq3$ \cite{azizian2020expressive}. As an example, k-order graph networks, introduced by \cite{morris2019weisfeiler}, offer expressiveness that is similar to a set-based variation of k-WL. \cite{maron2019provably} introduced a 2-order graph network that maintains expressive power similar to 3-WL. Furthermore, \cite{morris2020weisfeiler} introduced a localized variant of k-WL, focusing solely on a subset of vertices within a neighborhood. Nevertheless, using these expressive GNNs presents challenges due to their intrinsic computational demands and intricate architecture. In addition, some studies aimed to integrate inductive biases on isomorphism counting w.r.t predefined topological attributes such as triangles, cliques, and cycles \cite{bouritsas2020improving, liu2020neural, monti2018motifnet}. These efforts similar to the traditional graph kernels, as outlined by \cite{yanardag2015deep}. However, the task of predefining topological characteristics needs specialised knowledge in the respective domain, a resource that is frequently not easily accessible. (3) In a different vein, there has been a recent surge in studies exploring into the notion of enhancing GNNs through the augmenting of node identifiers or stochastic features. For example, \cite{vignac2020building} introduced an approach that preserves a node's local context through the manipulation of node identifiers in a permutation-equivariant fashion. \cite{you2021identity} developed ID-GNNs, incorporating vertex identity information in their design. \cite{chen2020can} and \cite{murphy2019relational} assigned one-hot identifiers to nodes, drawing inspiration from the principles of relational pooling. In a similar vein, \cite{sato2021random} enriched the representational capability of GNNs by incorporating a random feature for each node. There are some other approaches modify the MPNN framework or incorporate additional heuristics to enhance their expressiveness \cite{bouritsas2006improving, bodnar2021weisfeiler, wijesinghe2021new}. (4) Some works inject positional encoding (PE) as initial node features because nodes in a graph lack inherent positional information. Canonical index PE can be assigned to the nodes in a graph. However, the model must be trained on all possible index permutations, or sampling must be employed \cite{murphy2019relational}. Another direction for PE in graphs is using Laplacian Eigenvectors \cite{dwivedi2023benchmarking, dwivedi2020generalization}, as they establish a meaningful local coordinate system while maintaining the global structure of the graph. \cite{dwivedi2021graph} proposed a PE scheme (RWPE) based on random-walk diffusion to initialize the positional representations of nodes. These positional encoding methods such as Laplacian positional encoding \cite{dwivedi2020generalization} or RWPE \cite{dwivedi2021graph} have a significant limitation in that they usually fail to quantify the structural similarity between nodes and their surrounding neighborhoods. Nonetheless, while these techniques have demonstrated their expressivity to go beyond 1-WL. However, it remains uncertain what further attributes they can encompass beyond the scope of 1-WL. 

Despite these limitations, our method offers notable advantages. \emph{GenHopNet} enjoys greater expressive power than the 1-WL test, providing improved node and graph-level distinction by accounting for both local and global graph structures through closed walk counts and positional information. Additionally, by incorporating edge centrality measures to enrich message-passing, \emph{StructPosGSSL} enhances the model's ability to differentiate various types of connections, making it strictly more expressive than Subgraph MPNNs \cite{you2021identity, cotta2021reconstruction, zhang2021nested} in distinguishing certain non-isomorphic graphs.

\section{An Expressive and Generalizable $k$-hop Message Passing Framework}
In this section, we introduce the Expressive and Generalizable Message-Passing (EGMP) framework, designed to incorporate learnable local structural information through an aggregation method that leverages the $k$-hop neighborhood without the need for explicit extraction of local substructure patterns. We provide a theoretical analysis demonstrating how $k$-hop GNNs within this framework can achieve greater expressiveness than 1-WL.

Let $G = (V, E, \mathbf{A})$ be an undirected graph with a set of nodes $V$ and a set of edge $E$, where $|V|=m$, $|E|=e$, and $\mathbf{A} \in \mathbb{R}^{m\times m}$ is the adjacency matrix. 
Let $\mathbf{L}=\mathbf{D}-\mathbf{A}$ be a Laplacian matrix, where a diagonal matrix $\mathbf{D} \in \mathbb{R}^{m\times m}$, and ${D}_{ii}=\sum_{j}{A}_{ij}$. $\mathbf{L}$ is a real symmetric matrix diagonalizable as $\mathbf{L}=\mathbf{U} \mathbf{\Lambda} \mathbf{U}^{H}$. Moreover, $\mathbf{U}=\left\{{u_i}\right\}_{i=1}^{m} \in \mathbb{R}^{m}$ are orthogonal eigenvectors, $\mathbf{\Lambda}=\operatorname{diag}\left(\left[\lambda_1,\dots,\lambda_{m}\right]\right) \in \mathbb{R}^{m\times m}$ are real eigenvalues, and $\mathbf{U}^H$ is a hermitian transpose of $\mathbf{U}$.

Let $\{\!\!\{\cdot\}\!\!\}$ represent a multiset Let ${A}^k_{vu}\!=\!\big(\mathbf{A}^k\big)_{v,u}$ and 
define 
$\tilde{A}^k_{vu}\!=\!\frac{{A}^k_{vu}}{\sum_{u \in \mathcal{N}(v) \setminus v}{A}^k_{vu}}$ refers to a normalized value of ${A}^k_{vu}$, and $\mathbf{X} \in \mathbb{R}^{m \times d'}$ be the matrix of input node attributes with each $\mathbf{x}_v\in \mathbb{R}^{d'}$ corresponding to each vertex $v\in V$. We indicate the feature vector of vertex $v$ at the t\textsuperscript{th} layer as ${\mathbf{h}}_v^{(t)}$ and set $\mathbf{h}_v^{(0)}=\mathbf{x}_v$. Then, the definition of the (t+1)\textsuperscript{th} layer in our model is given as:
\begin{align}
&\mathbf{m}^{(t)}_{local\_pat}(v)  \!=\!\textsc{Agg}_{lp} \Big(\big\{\!\!\big\{({A}_{vu},\mathbf{h}^{(t)}_u,\mathbf{e}^{b}_{uv}, \mathbf{e}^{c}_{uv})| u \in \mathcal{N}(v) \big\}\!\!\big\}\Big),\!\!\label{eq:1}\\
&\mathbf{m}^{(t)}_{high\_ord}(v) \!=\!\textsc{Agg}_{ho} \Big(\big\{\!\!\big\{(\tilde{{A}}^k_{vu},\mathbf{h}^{(t)}_u )| u \in \mathcal{N}^k(v) \big\}\!\!\big\}\Big) ; k \ge 2, \label{eq:2}\\
&\mathbf{m}^{(t)}_{closed\_walks}(v) \!=\! \textsc{Agg}_{cw}\Big(\big\{\!\!\big\{{A}^k_{vv}, \mathbf{h}^{(t)}_v\big\}\!\!\}\Big) ; k \ge 2, \label{eq:3}\\
&\!\!\!\!\!\!\mathbf{h}^{(t+1)}_v\!\!=\!\textsc{Combine}\Big(\mathbf{h}^{(t)}_v, \mathbf{m}^{(t)}_{local\_pat}(v), \mathbf{m}^{(t)}_{high\_ord}(v), \mathbf{m}^{(t)}_{closed\_walks}(v)\Big).\label{eq:4}
\end{align}

\vspace{-0.2cm}
The above equations define a process for aggregating messages in our GNN. We define the following:
\renewcommand{\labelenumi}{\roman{enumi}.}
\begin{enumerate}[leftmargin=0.6cm]
    \item $\textsc{Agg}_{lp} (\cdot)$ in Eq. \ref{eq:1} aggregates the 1-hop neighborhood of vertex $v$ into $\mathbf{m}^{(t)}_{local\_pat}(v)$ based on the node embeddings $\mathbf{h}$, dataset edge attribute embeddings $\mathbf{e}^b$ and centrality edge attribute embeddings $\mathbf{e}^c$.
    \item $\textsc{Agg}_{ho}(\cdot)$ in Eq. \ref{eq:2} obtains a normalized $\mathbf{m}^{(t)}_{high\_ord}(v)$ from the $k$-hop neighbors ($k\ge 2$) of vertex $v$, weighting their contributions by coefficient $\tilde{{A}}^k_{vu}$ from the normalized adj. mat. $\tilde{\mathbf{A}}^k$. 
    \item $\textsc{Agg}_{cw}(\cdot)$ in Eq. \ref{eq:3} forms the closed-walk message  for vertex $v$, $\mathbf{m}^{(t)}_{closed\_walks}(v)$,  capturing closed-walks of length up to $k$ ($k \ge 2$) that depart from the node $v$ and return to the node $v$. The diagonal elements of $\mathbf{A}^k$, given as ${A}^k_{vv}$, count the number of closed walks of length $k$. 
    \item Eq. \ref{eq:4} combines the feature vector and all three types into the node representation of $v$ for the next layer.
\end{enumerate}

We use the above set of equations to compute the graph's topological information. For the positional information, we only use Eq. \ref{eq:1} by replacing  node features $\mathbf{h}^{(t)}_u$ with positional features $\mathbf{h}^{(t)}_{u, pos}$.

The mechanism in  Eq. \ref{eq:3} highlights the importance of node $v$ within its local topology and its role in the connectivity of the graph over multiple hops, exhibit the following properties:
\renewcommand{\labelenumi}{\arabic{enumi}.}
\begin{enumerate}[leftmargin=0.6cm]
    \item \textbf{Closed Walks and Connectivity.} The power $\mathbf{A}^k$  enumerates all possible walks of length $k$ in the graph. By examining ${A}^k_{vv}$, one can infer how \emph{connected} or \emph{central} a node is with respect to walks of length $k$.  $\textbf{Tr}\big(\mathbf{A}^k\big)$ counts the total number of closed walks of length $k$ starting and ending at the same vertices. Such a measure quantitfies the graph's connectivity: 
    {
    \renewcommand{\labelenumii}{\theenumii}
    \renewcommand{\theenumii}{\theenumi.\arabic{enumii}.}
\begin{enumerate}[leftmargin=0.55cm]
        \item \textbf{Local Connectivity.} High numbers of shorter closed walks (smaller $k$) indicate strong local connectivity. This is useful for understanding how tightly knit individual neighborhoods are within the graph.
        \item \textbf{Global Connectivity.} As $k$ increases, the nature of the closed walks provides insights into the global connectivity and the presence of cycles within the graph. Any cycle in a graph is  a closed walk. However, not all closed walks are cycles as cycles have the additional constraint of not repeating vertices/edges except the starting/ending vertex.
   \end{enumerate}
   }
    \item \textbf{Isomorphic Invariance.} Such a simple trick facilitates the creation of unique node representations by ensuring that each node possesses a distinct $k$-hop topological neighborhood, provided that $k$ is sufficiently large. The sum $\sum_{k} {A}^k_{vv}$ across different powers $k$ reflects the number of closed walks of varying lengths starting and ending at node $v$. For instance, the equality $\sum_{k} {A}^k_{vv} = \sum_{k} {A}^k_{v'v'}$ holds if vertices $v$ and $v'$ are $k$-hop isomorphic, implying that they share identical local connectivity patterns up to $k$-hops. This characteristic serves as a powerful tool for identifying and distinguishing nodes based on their structural roles within the network.
\end{enumerate}

\begin{figure}
    \centering
    \includegraphics[width=0.40\textwidth]{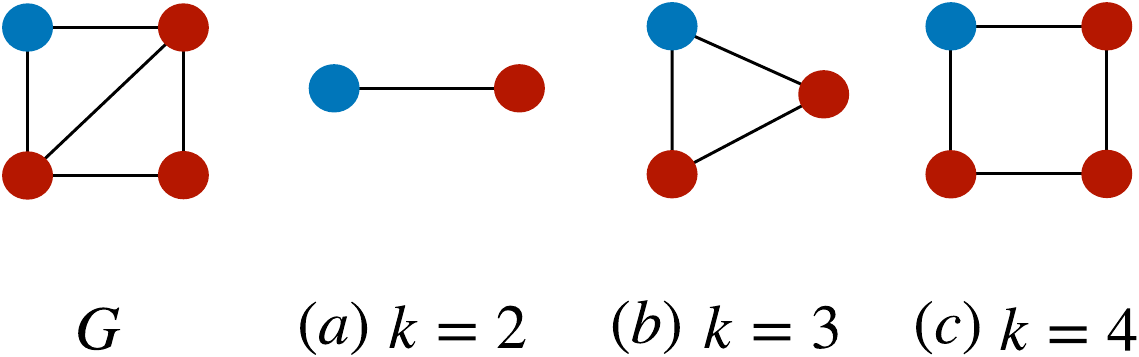}
    \caption{A high-level overview of $k=2, \dots, 4$ closed walks, where the blue node is the source node. For $k=2$, the walk can traverse between nodes multiple times, forming a non-cyclic path. For $k=3$ and $k=4$, the walks can capture closed cycles of length 3 and 4, respectively, effectively identifying cyclic structures in the graph.}
    \label{fig:closed-walk}
\end{figure}

Understanding cycle and closed-walk isomorphisms is essential for elucidating the structural roles of nodes within a graph, revealing both local and global connectivity patterns. The cycle isomorphism highlights nodes that engage in similar closed-loop interactions, while the closed-walk isomorphism provides insights into broader connectivity by capturing indirect relationships that contribute to the overall network topology. The definitions of the cycle and closed-walk isomorphisms formalize these concepts, emphasizing their significance in analyzing graph structures.

\begin{definition}
\textbf{Cycle Isomorphism ($\simeq_{Cycle})$}. Two nodes $v$ and $v'$ are cycle-isomorphic if the sum of the powers of $\mathbf{A}$ over $k$ (i.e., $\sum_{k} {A}^k_{vv}$ and $\sum_{k} {A}^k_{v'v'}$) considers only closed walks that are also cycles (i.e., walks that do not repeat any vertices or edges except the starting/ending vertex).
\end{definition}

\begin{definition}
\textbf{Closed Walk Isomorphism ($\simeq_{ClosedWalk}$)}. Two nodes $v$ and $v'$ are closed-walk isomorphic if the sum of the powers of $\mathbf{A}$ over $k$ (i.e., $\sum_{k} {A}^k_{vv}$ and $\sum_{k} {A}^k_{v'v'}$) considers all possible closed walks (i.e., walks that start and end at the same vertex, but may repeat vertices and edges). \vspace{-0.3cm}
\end{definition}

\noindent\begin{theorem}\label{theorem-1} The following statement is true: (a) If $\sum_{k} \mathbf{A}^k_{vv} \simeq_{Cycle} \sum_{k} \mathbf{A}^k_{v'v'}$, then $\sum_{k} \mathbf{A}^k_{vv} \simeq_{ClosedWalk} \sum_{k} \mathbf{A}^k_{v'v'}$; but not vice versa.
\end{theorem}

Going back ot Eq. \ref{eq:1},  $\mathbf{m}^{(t)}_{local\_pat}$ is a message aggregation function that considers the 1-hop neighborhood, including their MLP embeddings of the original edge attributes and the edge centrality attributes. We enrich the 1-hop neighborhood aggregation by injecting MLP embeddings of three different centrality based edge attributes 
between given pair of nodes: (1) Edge Betweenness (EB), (2) Edge Closeness (EC), and (3) Edge Clustering Coefficients (ECC)   \cite{wang2011identification}. Such attributes are designed to improve the distinguishing capabilities of the model, helping it  differentiate between various types of connections, and facilitating  nuanced understanding and processing of edge-related information in graphs. The centrality measures exhibit the following properties:
\renewcommand{\labelenumi}{\arabic{enumi}.}
\begin{enumerate}[leftmargin=0.6cm]
    \item \textbf{Local Connectivity.} Both EC and ECC provide insights into the local structure around an edge, highlighting how embedded an edge is within its immediate neighborhood and how it contributes to local connectivity and cohesiveness.
    \item \textbf{Global Connectivity.} EB extends to more global properties, reflecting the strategic importance of an edge across the entire network. It helps in understanding the potential vulnerabilities of the network, identifying crucial links whose removal or failure might significantly disrupt network connectivity.
    \item \textbf{Isomorphic Invariance.} All three measures (EB, EC, and ECC) are isomorphic-invariant measures as they are  based on the relationships and distances between nodes, which are preserved under graph isomorphism. 
\end{enumerate}

The feature vector for the next layer $\mathbf{h}^{(t+1)}_{v}$, is derived by combining representations as per Eq. \ref{eq:4}.

\section{Generalizable $k$-hop Network}
Below, we present the GNN model design based on the EGMP framework that we proposed in the previous section. 
We introduce a new GNN model called \emph{GenHopNet} (Generalizable $k$-hop Network), which utilizes an aggregation scheme based on our generalized message-passing framework. We demonstrate that the expressive power of \emph{GenHopNet} exceeds those of the 1-WL. For each vertex $v \in V$, the feature vector for the $(t+1)$\textsuperscript{th} layer is produced by:
\begin{align}
&\mathbf{h}^{(t+1)}_v = \textsc{Mlp}_{\phi} \bigg[(1+\epsilon) \,\mathbf{h}^{(t)}_v + \underbrace{\sum\nolimits_{u \in \mathcal{N}(v)}\!{A}_{vu} \Big(\mathbf{h}^{(t)}_u + \mathbf{e}^{b}_{vu} + \mathbf{e}^{c}_{vu}\Big)}_{\mathbf{m}^{(t)}_{local\_pat}}\nonumber\\[-20pt]
&\label{eq:gen_hop_net}\\
& \qquad\qquad\quad+\underbrace{\sum\nolimits_{k=2}^{K} {A}^{k}_{vv} \mathbf{h}^{(t)}_v}_{\mathbf{m}^{(t)}_{closed\_walks}} +\underbrace{\sum\nolimits_{k=2}^{K} \sum\nolimits_{u \in \mathcal{N}^{k}(v)} \Tilde{{A}}^{k}_{vu} \mathbf{h}^{(t)}_u}_{\mathbf{m}^{(t)}_{high\_ord}} \bigg],\nonumber
\end{align}
where $\epsilon$ is a learnable scalar.  

Moreover, the graph-level embedding $\mathbf{z}$ is computed using a sum pooling function over the final node embeddings as follows:
\begin{equation}\label{eq:gen_hop_net2}
    \mathbf{z} = \textsc{Mlp}_{\theta} \Big[\sum\nolimits_{v \in G} \mathbf{h}^{(t)}_v\Big].
\end{equation}
Figure \ref{fig_1_1} summarizes our pipeline inclusive of Eq. \ref{eq:gen_hop_net} and \ref{eq:gen_hop_net2}.

\vspace{0.1cm}
\noindent\textbf{Expressiveness analysis.} We first generalize the result of universal functions over \emph{multisets} \cite{xu2018powerful} to universal functions over \emph{pairs of multisets} since Eq.~\ref{eq:gen_hop_net} involves not only node features but also edge features $\mathbf{e}_{vu}^b$, centrality based edge features $\mathbf{e}_{vu}^c$, normalized $k$-hop neighboring coefficients $\Tilde{{A}}^k_{uv}$ and $k$-hop closed-walk coefficients ${A}^k_{vv}$. Let $\mathcal{H}$, $\mathcal{A}$, $\mathcal{A}^k$, $\Tilde{\mathcal{A}}^k$, $\mathcal{W}_1$, $\mathcal{W}_2$, and $\mathcal{W}_3$ be countable sets where $\mathcal{H}$ is a node feature space, $\mathcal{A}$ is a 1-hop neighborhood coefficient space, $\mathcal{A}^k$ is a $k$-hop closed-walk coefficient space, $\Tilde{\mathcal{A}}^k$ is a normalized $k$-hop neighborhood coefficient space. Moreover, $\mathcal{W}_1\!=\!\big\{{A}_{vu}(\mathbf{h}_u \!+\! \mathbf{e}_{vu}^b \!+\! \mathbf{e}_{vu}^c)|{A}_{vu}\!\in\! \mathcal{A},\; \mathbf{h}_u\in \mathcal{H}, \mathbf{e}_{vu}^b \in \mathcal{E}^b, \mathbf{e}_{vu}^c \in \mathcal{E}^c\big\}$,\; $\mathcal{W}_2\!=\!\big\{\Tilde{{A}}^k_{vu} \mathbf{h}_u |\Tilde{{A}}^k_{vu}\!\in\! \Tilde{\mathcal{A}}^k, \;\mathbf{h}_u\in \mathcal{H}\big\}$, and $\mathcal{W}_3\!=\!\big\{{A}^k_{vv}\mathbf{h}_v |{A}^k_{vv}\!\in\! {\mathcal{A}}^k, \mathbf{h}_v\in \mathcal{H}\big\}$.

The following theorem asserts that a GNN can surpass the expressiveness of 1-WL provided that our framework is sufficiently robust to differentiate the structures beyond neighborhood subtrees, and the neighborhood aggregation function is injective, given a sufficient number of hops $k > 1$.

\begin{theorem}\label{theorem-EGMP} Let $S$ represent a GNN with an aggregation scheme $\pi'(\cdot)$ delineated by Eq. \ref{eq:1}\ref{eq:4}. $S$ exceeds the expressiveness of 1-WL in identifying non-isomorphic graphs, provided that $S$ operates over a sufficient number of hops, $k > 1$, and also meets the following criteria:
\begin{itemize}
    \item[(1)] $\pi' \Big(\mathbf{h}^{(t)}_v, \{\!\!\big\{({A}_{vu}, \mathbf{h}^{(t)}_u, \mathbf{e}^{b}_{uv}, \mathbf{e}^{c}_{uv})| u \in \mathcal{N}(v) \big\}\!\!\}, \{\!\!\big\{(\tilde{A}^k_{vu}, \mathbf{h}^{(t)}_u )| u \in \mathcal{N}^k(v) \big\}\!\!\big\}, \{\!\!\big\{({A}^k_{vv}, \mathbf{h}^{(t)}_v\big)\}\!\!\}\Big)$ is injective (Eq. ~\ref{eq:gen_hop_net});
    \item[(2)] The graph-level readout function of $S$ is injective (Eq. ~\ref{eq:gen_hop_net2}). \vspace{-0.2cm}
\end{itemize}
\end{theorem}

\begin{figure}
    \centering
    \includegraphics[width=0.5\textwidth] {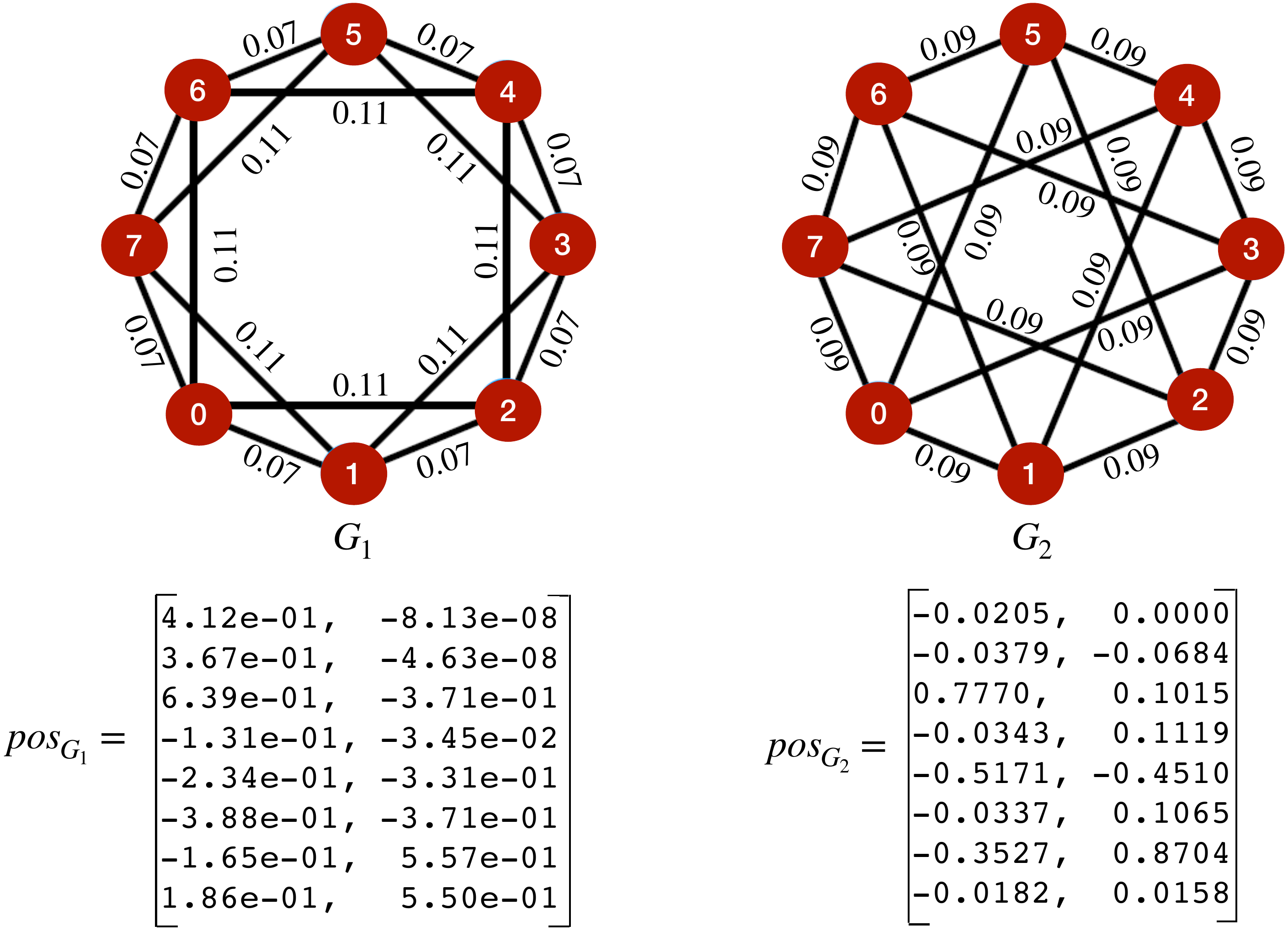}
    \caption{A pair of non-isomorphic graphs where positional encoding (with a dimension of 2) and EB attributes outperform the 1-WL test in distinguishing between graphs $G_1$ and $G_2$, allowing for the detection of structural differences that the 1-WL test fails to capture.}
    \label{fig:closed-walk-eb-pairs}
\end{figure}

\begin{lemma}\label{lem:lem-1}
Given two distinct pairs of multisets $\mathbf{W}_1, \mathbf{W}_1^{'} \in \mathcal{W}_1$, $\mathbf{W}_2, \mathbf{W}_2^{'} \in \mathcal{W}_2$, $\mathbf{W}_3, \mathbf{W}_3^{'} \in \mathcal{W}_3$, there exists a function $f$ such that the aggregation function $\pi(\mathbf{W}_1, \mathbf{W}_2, \mathbf{W}_3)$ and $\pi(\mathbf{W}_1^{'}, \mathbf{W}_2^{'}, \mathbf{W}_3^{'})$ defined as $\pi(\mathbf{W}_1, \mathbf{W}_2, \mathbf{W}_3)=\sum_{\mathbf{w}_1 \in \mathbf{W}_1} f(\mathbf{w}_1)+\sum_k \Big(f(\mathbf{w}_3)+\sum_{\mathbf{w}_2 \in \mathbf{W}_2} f(\mathbf{w}_2)\Big)$ and $\pi(\mathbf{W}_1^{'}, \mathbf{W}_2^{'}, \mathbf{W}_3^{'})=\sum_{\mathbf{w}_1^{'} \in \mathbf{W}_1^{'}} f(\mathbf{w}_1^{'})+\sum_k \Big(f(\mathbf{w}_3^{'})+\sum_{\mathbf{w}_2^{'} \in \mathbf{W}_2^{'}} f(\mathbf{w}_2^{'})\Big)$ are unique, respectively.
\end{lemma}

\begin{lemma}\label{lem:lem-2}
Expanding upon Lemma 1, we introduce an extended aggregation function $\pi'(\mathbf{h}_v, \mathbf{W}_1, \mathbf{W}_2, \mathbf{W}_3)$, which incorporates the feature vector of the central node $\mathbf{h}_v$ and the multisets $\mathbf{W}_1 \in \mathcal{W}_1$, $\mathbf{W}_2 \in \mathcal{W}_2$, and $\mathbf{W}_3 \in \mathcal{W}_3$. There exists a function $f$ such that $\pi'(\mathbf{h}_v, \mathbf{W}_1, \mathbf{W}_2, \mathbf{W}_3)=(1+\epsilon) f(\mathbf{h}_v)+\sum_{\mathbf{w}_1 \in \mathbf{W}_1} f(\mathbf{w}_1)+\sum_k \Big(f(\mathbf{w}_3)+\sum_{\mathbf{w}_2 \in \mathbf{W}_2} f(\mathbf{w}_2)\Big)$ is unique for any distinct quadruple $(\mathbf{h}_v, \mathbf{W}_1, \mathbf{W}_2, \mathbf{W}_3)$, where $\mathbf{h}_v\in \mathcal{H}$, $\mathbf{w}_3 \in \mathbf{W}_3$, and $\epsilon$ is an arbitrary real number.
\end{lemma}

\begin{corollary}
GenHopNet exhibits greater expressiveness compared to 1-WL when evaluating non-isomorphic graphs.
\end{corollary}

Our proposed graph neural network model $S$, which operates on up to $k$-hop  information for feature aggregation, surpasses the expressiveness of 1-WL in distinguishing non-isomorphic graphs. This is achieved by ensuring the uniqueness of feature representations through the extended aggregation function $\pi'(\mathbf{h}_v, \mathbf{W}_1, \mathbf{W}_2, \mathbf{W}_3)$, as established by Lemma 2. Thus, $S$ can capture and differentiate structural nuances beyond what 1-WL can achieve, making it a powerful tool for graph classification tasks.

\vspace{0.1cm}
\noindent
\textbf{Complexity Analysis.} Similar to GIN \cite{hou2019measuring}, \emph{GenHopNet} is computationally efficient, with its time and memory complexities scaling linearly in relation to the number of edges in the graph. The time and space complexities of \emph{GenHopNet} are $\mathcal{O}(ted'd)$ and $\mathcal{O}(e)$, respectively, where $e$ denotes the number of edges in the graph, $t$ represents the number of layers, and $d'$ and $d$ correspond to the dimensions of the input and output feature vectors.

\section{Graph Self-Supervised Learning Framework}
Below, we introduce Structural and Positional GSSL (\emph{StructPosGSSL}), a new class of graph self-supervised learning framework based on structural and positional information within graphs.

\subsection{Data Augmentation for Graph}
The goal of data augmentation is to produce consistent, identity-preserving positive samples of a specific graph. In this work, we use two main types of augmentation strategies: structural augmentation and feature augmentation \cite{you2020graph}. In structural augmentation, three distinct strategies are considered: (1) Subgraph Induction by Random Walks (RWS), (2) Node Dropping (ND), and (3) Edge Dropping (ED). For feature augmentation, we employ three different approaches: (1) Feature Dropout (FD), (2) Feature Masking (FM), and (3) Edge Attribute Masking (EAM). In our work, we generate different augmented graphs from a single input graph. 
One view uses augmented attributes of nodes (or edges) and the other view uses topologically augmented graph.

\subsection{Expressive Graph Encoders}
For each augmented view, we process it through two distinct GNN encoders: (i) the structural encoder (the proposed \emph{GenHopNet}) snd (ii) the positional encoder. We initiate the positional feature vectors using the graph's Laplacian eigenvectors, as outlined in \cite{dwivedi2020generalization}. This second encoder, also a GNN, applies Eq.\ref{eq:1} with the $\textsc{Combine}$ function, \ie, $\mathbf{h}^{(t+1)}_{v, pos}  =\textsc{Combine}\Big(\mathbf{h}^{(t)}_{v, pos}, \mathbf{m}^{(t)}_{local\_pat}(v)\Big)$, and leverages the spectral properties of the graph Laplacian. 

By focusing on the spectral characteristics of the Laplacian, this encoding strategy effectively captures both the local connectivity of nodes and the broader topology of the graph, and the node's ``position'' in that topology, significantly enhancing the model's capability to understand and manage complex graph structures.

Each encoder (structural and positional) outputs node representations and a final graph representation for augmented views. We then pass them through another shared projection head (MLP) to obtain the final structural and positional representations for both nodes and graphs. Next, we concatenate the structural features with positional features for node representations and graph representations separately, ensuring a comprehensive integration of both structural and positional data. To facilitate the end-to-end training of encoders and generate comprehensive node and graph representations that are independent of specific downstream tasks, we employ 
the graph- and node-wise contrastive loss functions.

\subsection{Training Pipeline}

To enable end-to-end training of the encoders and to develop robust graph and node representations that are independent of downstream tasks, 
we employ the NT-Xent \cite{chen2020simple} loss to learn the graph representations and the  VICReg \cite{bardes2022vicreg} loss to learn the node representations. The NT-Xent loss maximizes discriminative power between positive and negative samples of graph representations. The VICReg  ensures a balanced and non-redundant spread of node features across embedding dimensions to reduce the representational collapse. This integration  stabilizes training in self-supervised setup and enriches   graph representations for downstream tasks.

The NT-Xent loss is defined as:
\begin{equation}\label{equ:nt-xent}
L_{graph} (\mathbf{z}^i, \mathbf{z}^j) = -log\, \frac{exp\big(sim(\mathbf{z}^i, \mathbf{z}^j) / \tau\big)}{\sum_{k=1}^{2n} \textbf{1}_{[k \neq i]}exp\big(sim(\mathbf{z}^{i}, \mathbf{z}^k) / \tau\big)},
\end{equation}
where $\mathbf{z}^{i}, \mathbf{z}^{j} \in \mathbb{R}^{d^\dagger}$ are graph representations for two augmented views,  $d^\dagger$ is the feature size after concatenation of structural and positional embeddings, $\tau$ is the so-called  temperature, and $n$ is the number of samples in the batch (total is $2n$ after applying augmentations). Moreover, $\textbf{1}_{[k \neq i]}$ is an indicator function that is $\textbf{1}$ if $k \neq i$ and $0$ otherwise, whereas $sim(\cdot, \cdot)$ is the cosine similarity. 

\definecolor{LightCyan}{rgb}{0.88,1,0.88}
\begin{table}
\centering 
\scalebox{0.95}{\begin{tabular}{l c c} 
\specialrule{.1em}{.05em}{.05em}
Method & CSL & SR25 \\ [0.5ex] 
\hline 
GCN & 10.0 $\pm$ 0.0 & 6.6 $\pm$ 0.0 \\
GIN &  10.0 $\pm$ 0.0 & 6.6 $\pm$ 0.0 \\ 
3WLGNN & 97.8 $\pm$ 10.9 & - \\ 
3-GCN & 95.7 $\pm$ 14.8 & 6.6 $\pm$ 0.0 \\
GCN-RNI & 16.0 $\pm$ 0.0 & 6.6 $\pm$ 0.0 \\
\hline
\rowcolor{LightCyan}StructPosGSSL-SA & \textbf{98.6} $\pm$ \textbf{2.8} & \textbf{100.0} $\pm$ \textbf{0.0} \\ 
\rowcolor{LightCyan}StructPosGSSL-FA & \textbf{98.3} $\pm$ \textbf{2.5} & \textbf{100.0} $\pm$ \textbf{0.0} \\ 
\specialrule{.1em}{.05em}{.05em}
\end{tabular}}
\caption{Class. acc. (\%) on the test set for CSL/SR25 datasets. 
\label{Tab:experiment_expressive}}
\end{table}

Next, we combine the refined VICReg loss with the contrastive NT-Xent loss to create a unified loss function that maximizes mutual information between graph embeddings while preserving structural and positional node alignment. This unified approach not only aligns embeddings for isomorphic node pairs but also maintains diversity and enhances the model's topological discriminative power. By enabling the model to differentiate between nodes with subtle structural or positional differences, this method is crucial for accurately identifying both isomorphic and non-isomorphic node pairs. In this work, the  VICReg loss is  adapted for correspondence node alignment to improve isomorphic graph representation learning in a self-supervised setting.

The VICReg loss is given as:
\begin{equation}\label{equ:vicreg}
\begin{split}
    L_{node}(\mathbf{H}^{i}, \mathbf{H}^{j}) &= \lambda_{inv} \, L_{inv}\big(\mathbf{H}^{i}, \mathbf{H}^{j}\big) + \lambda_{var} \, L_{var}\big(\mathbf{H}^{i}, \mathbf{H}^{j}\big) + \\& \qquad\qquad\;\lambda_{cov} \, L_{cov}\big(\mathbf{H}^{i}, \mathbf{H}^{j}\big),
\end{split}
\end{equation}
where $\lambda_{inv}, \lambda_{var}$ and $\lambda_{cov}$ are weighting factors for the invariance, variance, and covariance components of VICReg. The invariance loss $L_{inv}(\mathbf{H}^{i}, \mathbf{H}^{j})=\frac{1}{m} \sum_{v} \big\lVert\mathbf{h}^{i}_v-\mathbf{h}^{j}_v\big\rVert_2^2$, where $\mathbf{h}^{i}, \mathbf{h}^{j} \in \mathbb{R}^{d}$ are node representations for two augmented views. The var. loss $L_{var}(\mathbf{H}^{i}, \mathbf{H}^{j})\!=\!\frac{1}{d} \!\sum\limits_{l=1}^{d}\!\bigg\lvert\max\Big(0, \gamma\! -\! \operatorname{std}\big(\mathbf{H}^{i}_{l,:}, \epsilon\big)\Big)\!-\!\max\Big(0, \gamma \!-\! \operatorname{std}\big(\mathbf{H}^{j}_{l,:}, \epsilon\big)\Big)\bigg\rvert$. The cov. loss is $L_{cov}(\mathbf{H}^{i}, \mathbf{H}^{j})=\frac{1}{d} \sum\limits_{l \neq q} \bigg\lvert{\Big({C}^i_{l, q}\Big)}^2 - {\Big({C}^j_{l, q}\Big)}^2\bigg\rvert$ where $\textbf{C}=\frac{1}{m-1} \sum_{l=1}^{m} (\mathbf{h}_l - \bar{\mathbf{h}}) (\mathbf{h}_l - \bar{\mathbf{h}})^\top$ and $\bar{\mathbf{h}}=\frac{1}{m} \sum_{l=1}^{m} \mathbf{h}_l$. The invariance loss captures the inherent structure of the graph, maintaining node representations' structural and positional properties despite augmentation. The variance loss helps distinguish non-isomorphic nodes, enhancing the model's ability to capture structural differences. Lastly, the covariance loss ensures that node embeddings remain spread in the embedding space.

The total combined loss for integrating NT-Xent and refined VICReg is  the sum of these two losses:
\begin{equation}\label{equ:loss}
\begin{split}
    L_{total}\big(\mathbf{z}^{i}, \mathbf{z}^{j}, \mathbf{H}^{i}, \mathbf{H}^{j}\big) = L_{graph} \big(\mathbf{z}^{i}, \mathbf{z}^{j}\big) + \alpha \, L_{node}\big(\mathbf{H}^{i}, \mathbf{H}^{j}\big), 
\end{split}
\end{equation}
where $\alpha$ is the weighting factor for VICReg loss. 

\noindent\begin{theorem}\label{theorem-4} The {StructPosGSSL} is more expressive than subgraph MPNNs in distinguishing certain non-isomorphic graphs.
\end{theorem}


\section{Numerical Experiments}
\label{sec:experiments}
In this section, we evaluate our self-supervised learning framework on graph classification benchmark tasks. The results from our models are statistically significant with a 95\% confidence level. 
We evaluate \emph{StructPosGSSL} on graph classification benchmark tasks and compare their performance with leading baselines to address the following questions:
\renewcommand{\labelenumi}{Q\arabic{enumi}.}
\begin{enumerate}[leftmargin=0.75cm]
    \item How effective are \emph{StructPosGSSL} in small graph classification tasks based on empirical performance?
    \item How effective are \emph{StructPosGSSL} in large graph classification tasks based on empirical performance?
    \item How effective is \emph{StructPosGSSL} for isomorphism testing in synthetic graph classification tasks?
    \item How do structural and positional encodings impact the overall performance? 
\end{enumerate}
In the following sections, we analyze the experimental results to address the four above questions.

\subsection{Experiments on Small Graphs}\label{sec:small-graphs}
We use eight datasets from two categories: 

\noindent
(1) bioinformatics datasets: MUTAG, PTC-MR, NCI1, and PROTEINS \cite{debnath1991structure, kriege2016valid, wale2008comparison, shervashidze2011weisfeiler}; (2) social network datasets: IMDB-B, IMDB-M, COLLAB and RDT-M5K \cite{yanardag2015deep}. 
We compare our method against fourteen baseline approaches:

\renewcommand{\labelenumi}{(\arabic{enumi})}
\begin{enumerate}[leftmargin=0.6cm]
\item Graph kernel methods: WL-OA \cite{kriege2016valid}, RetGK \cite{zhang2018retgk}, P-WL \cite{rieck2019persistent}, and WL-PM \cite{nikolentzos2017matching};
\item GNN-based methods: PATCHY-SAN \cite{niepert2016learning}, CAPSGNN \cite{xinyi2018capsule}, GIN \cite{xu2018powerful}, and G3N \cite{wang2023mathscr};
\item Unsupervised methods: GraphCL \cite{you2020graph}, MVGRL \cite{hassani2020contrastive}, GCS \cite{wei2023boosting}, GALOPA \cite{wang2024galopa}, GA2C \cite{liu2024graph} and GraphCL+HTML \cite{li2024hierarchical}.
\end{enumerate}


Specifically, the \emph{GenHopNet} encoder models are initially trained in an unsupervised manner, and the resulting embeddings are then passed into a linear classifier to accommodate the labeled data. Then, for fair comparison, we execute our method using ten random splits \cite{zhu2021empirical} and apply  10-fold cross-validation  and present the best mean accuracy (\%) along with the standard deviation. The results are presented in table \ref{Tab:bio-graph-baselines} and \ref{Tab:social-graph-baselines}. We have two settings: (1) StructPosGSSL-SA, which considers structure augmentation, and (2) StructPosGSSL-FA, which considers feature augmentation. In both settings, we employ the Adam optimizer \cite{kingma2014adam}, with hidden dimension of 128, weight decay is 0.0003, a 2-layer MLP with batch normalization, 100 epochs, positional encoding dimension of 6, a dropout rate of 0.5, and a temperature scaling parameter $\tau$ of 0.10. We choose a batch size from $\{32, 64, 128, 256\}$ and a number of hops $k \in \{2, 3, 4, 5, 6 \}$. We use $\lambda_{inv}=1, \lambda_{var}=3, \lambda_{cov}=2$, and $\alpha=0.04$ for MUTAG and $\lambda_{inv}=1, \lambda_{var}=25, \lambda_{cov}=25$, and $\alpha=0.005$ for PTC-MR, and $\lambda_{inv}=1, \lambda_{var}=24, \lambda_{cov}=24$, and $\alpha=0.005$ for the remaining datasets. The readout function, as described in \cite{xu2018powerful}, is utilized, which involves concatenating representations from all layers to derive a final graph representation.

To address \textbf{Q1}, in Tables \ref{Tab:bio-graph-baselines} \&  \ref{Tab:social-graph-baselines}, StructPosGSSL outperforms the best baseline  by $0.4 \%$ (PATCHY-SAN), $1.1 \%$ (CapsGNN), $1.1 \%$ (WL-OA, MVGRL), $0.5 \%$ (WL-PM), and $0.2 \%$ (GIN, GCS) on the datasets MUTAG, PTC-MR, IMDB-B, IMDB-M, and RDTM5K, respectvely.

These gains are a reflection of the inherent characteristics of the datasets. Graphs with smaller diameters, \ie, IMDB-B, IMDB-M, PTC-MR, and MUTAG, feature nodes that are closer together, promoting localized interactions that enable GNNs to capture both local and global information effectively, even with few message-passing steps. In such cases, structural encoding with closed walks is particularly beneficial, as it differentiates local structures by capturing repeated node interactions and identifying cycles. Conversely, datasets such as NCI1 and COLLAB, with larger diameters, present increased structural complexity, making it  challenging to capture patterns using closed walks alone due to the difficulty of accounting for distant node interactions with limited local information.

\subsection{Experiments on Large Graphs}
We use five large graph datasets from the Open Graph Benchmark (OGB) \cite{hu2020open}, comprising one molecular graph dataset (ogbg-moltoxcast, ogbg-moltox21, ogbg-molhiv, ogbg-molpcba) and one protein-protein association network (ogbg-ppa). We compare our approach with the  methods that provide results on the aforementioned OGB datasets: GIN+VN \cite{hu2020open}, GSN \cite{bouritsas2022improving}, GraphSNN \cite{wijesinghe2021new}, DS-GNN (EGO+), DSS-GNN (EGO+) \cite{bevilacquaequivariant}, and POLICY-LEARN \cite{bevilacquaefficient}.

For large graph datasets, we adopt the same experimental framework as outlined by Hu \etal \cite{hu2020open}. Our evaluation process is divided into two distinct learning phases. In the initial phase, the models are trained in a self-supervised fashion using only node features and graph structure without any label data. Subsequently, in the second phase, the representations generated by the GNN encoders during the first phase are fixed in place and employed to train, validate, and test the models using a straightforward linear classifier using the 10-fold cross-validation method.

We utilize the Adam optimizer with a learning rate of 0.001, a batch size of 32, dropout of 0.5, positional encoding dimension of 6, and run training for 100 epochs across all datasets. We use a 2-layer MLP with a hidden dimension of 200 and a temperature scaling parameter $\tau$ of 0.10 for both settings. We choose $\lambda_{inv}=1, \lambda_{var}=24, \lambda_{cov}=24$, and $\alpha=0.005$ for all datasets. The classification accuracy results are presented in Table \ref{Tab:experiment_OGB}.

To address \textbf{Q2}, in Table \ref{Tab:experiment_OGB}, StructPosGSSL consistently outperforms all the baseline methods across all OGB graphs listed. StructPosGSSL surpasses  best results of existing GNNs by $0.50 \%$ (POLICY-LEARN), $0.55 \%$ (DSS-GNN (EGO+)), $0.32 \%$ (GIN+VN), $0.24 \%$ (GraphSNN), and $0.42 \%$ (GIN+VN) on the datasets ogbg-moltoxcast, ogbg-moltox21, ogbg-molhiv, ogbg-ppa, and ogbg-molpcba, respectively.

\begin{table}
\centering 
\scalebox{0.90}{\hspace{-0.2cm}\begin{tabular}{r l c c c c c c}
& Method & MUTAG & PTC-MR & NCI1 & PROTEINS \\ [0.5ex] 
\hline
\multirow{4}{*}{\parbox{3cm}{\centering Graph kernel methods}}
& WL-OA & 84.5 $\pm$ 1.7 & 63.6 $\pm$ 1.5 & 86.1 $\pm$ 0.2 & 74.2 $\pm$ 0.4 \\
& RetGK & 90.3 $\pm$ 1.1 & 62.5 $\pm$ 1.6 & 84.5 $\pm$ 0.2 & 72.3 $\pm$ 0.6 \\
& P-WL & 90.5 $\pm$ 1.3 & 64.0 $\pm$ 0.8 & 85.4 $\pm$ 0.1 & 70.4 $\pm$ 0.1 \\
& WL-PM & 87.7 $\pm$ 0.8 & 61.4 $\pm$ 0.8 & \textbf{86.4} $\pm$ \textbf{0.2} & 73.6 $\pm$ 0.2 \\
[0.5ex]\hline
\multirow{4}{*}{\parbox{3cm}{GNN-based methods}} & PATCHY-SAN & 92.6 $\pm$ 4.2 & 60.0 $\pm$ 4.8 & 78.6 $\pm$ 1.9 & 73.1 $\pm$ 2.4 & \\ 
& CapsGNN & 86.6 $\pm$ 1.5 & 66.0 $\pm$ 1.8 & 78.3 $\pm$ 1.3 & 73.1 $\pm$ 4.8 \\
& GIN* & 89.4 $\pm$ 5.6 & 64.6 $\pm$ 7.0 & 82.7 $\pm$ 1.7 & 
\textbf{75.1} $\pm$ \textbf{5.1} \\
& G3N & 89.9 $\pm$ 8.0 & - & 82.3 $\pm$ 5.2 & 73.6 $\pm$ 5.3
\\[0.5ex]\hline
\multirow{6}{*}{\parbox{3cm}{\centering Unsupervised methods}} 
& GraphCL & 86.8 $\pm$ 1.3 & 61.3 $\pm$ 2.1 & 77.9 $\pm$ 0.4 & 74.4 $\pm$ 0.4 \\
& MVGRL & 89.7 $\pm$ 1.1 &  62.5 $\pm$ 1.7 & 77.0 $\pm$ 0.8 & - \\
& GraphCL+HTML & 88.9 $\pm$ 1.8	& - & 78.7 $\pm$ 0.7 & 74.9 $\pm$ 0.3 \\
& GCS & 90.4 $\pm$ 0.8 & - & 77.3 $\pm$ 0.3 & 75.0 $\pm$ 0.3 \\
& GALOPA & 91.1 $\pm$ 1.2 & - & 77.8 $\pm$ 0.3 & \textbf{76.9} $\pm$ \textbf{0.1} \\
& GA2C & 90.3 $\pm$ 0.3 & - & 80.6 $\pm$ 0.3 & 76.0 $\pm$ 0.1 \\
[0.5ex]\hline
\rowcolor{LightCyan}& StructPosGSSL-{SA} & \textbf{93.0} $\pm$ \textbf{5.3} & \textbf{67.1} $\pm$ \textbf{4.9} & 82.9 $\pm$ 3.8 & 75.8 $\pm$ 2.6 \\
\rowcolor{LightCyan}\multirow{-2}{*}{\parbox{2cm}{Our work}}& StructPosGSSL-{FA} & 91.5 $\pm$ 2.5 & 66.5 $\pm$ 3.7 & 82.5 $\pm$ 3.5 & 75.3 $\pm$ 2.9 \\
\specialrule{.1em}{.05em}{.05em}
\end{tabular}}\caption{Classification accuracy (\%) on bioinformatics datasets averaged over 10 runs.} 
\label{Tab:bio-graph-baselines}
\end{table}

\begin{table}
\centering 
\scalebox{0.90}{\hspace{-0.2cm}\begin{tabular}{r l c c c c} 
& Method & IMDB-B & IMDB-M & COLLAB & RDTM5K \\ [0.5ex] 
\hline
\multirow{4}{*}{\parbox{3cm}{\centering Graph kernel methods}}
& WL-OA & 74.2 $\pm$ 0.4 & 51.3 $\pm$ 0.2 & 80.7 $\pm$ 0.1 & - \\
& RetGK & 72.3 $\pm$ 0.6 & 48.7 $\pm$ 0.6 & \textbf{81.0} $\pm$ \textbf{0.3} & 56.1 $\pm$ 0.5 \\
& P-WL & 70.4 $\pm$ 0.1 & 50.9 $\pm$ 0.3 & - & - \\
& WL-PM &  73.6 $\pm$ 0.2 & 52.3 $\pm$ 0.2 & 81.5 $\pm$ 0.5 & - \\
[0.5ex]\hline
\multirow{4}{*}{\parbox{3cm}{GNN-based methods}} & PATCHY-SAN & 73.1 $\pm$ 2.4 & - & 72.6 $\pm$ 2.2 & 49.1 $\pm$ 0.7 \\ 
& CapsGNN & 73.1 $\pm$ 4.8 & 51.1 $\pm$ 3.1 & 79.6 $\pm$ 2.9 & 52.8 $\pm$ 1.4 \\
& GIN* & 75.1 $\pm$ 5.1 & 52.3 $\pm$ 2.8 & 80.2 $\pm$ 1.9 & 57.0 $\pm$ 1.7 \\
& G3N & 73.6 $\pm$ 3.5 & - & - & - 
\\[0.5ex]\hline
\multirow{6}{*}{\parbox{3cm}{\centering Unsupervised methods}} 
& GraphCL & 71.1 $\pm$ 0.4 &  49.2 $\pm$ 0.6 & 71.4 $\pm$ 1.2 & 55.9 $\pm$ 0.2 \\
& MVGRL & 74.2 $\pm$ 0.7 & 51.2 $\pm$ 0.5 & 76.0 $\pm$ 1.2 & - \\
& GraphCL+HTML & 71.6 $\pm$ 0.4	& - & - & 55.9 $\pm$ 0.3 \\
& GCS & 73.4 $\pm$ 0.3 & - & - & 57.0 $\pm$ 0.4 \\
& GALOPA & 70.7 $\pm$ 0.4 & - & - & - \\
& GA2C & 73.8 $\pm$ 0.3 & - & - & 55.9 $\pm$ 0.6 \\
[0.5ex]\hline
\rowcolor{LightCyan}& StructPosGSSL-{SA} &  \textbf{75.5} $\pm$ \textbf{3.3} & \textbf{52.8} $\pm$ \textbf{3.5} & 76.5 $\pm$ 3.6 & \textbf{57.2} $\pm$ \textbf{3.3} \\
\rowcolor{LightCyan}\multirow{-2}{*}{\parbox{2cm}{Our work}}& StructPosGSSL-{FA} & 75.1 $\pm$ 3.1 & 52.3 $\pm$ 3.6 & 76.1 $\pm$ 3.3 & \textbf{56.8} $\pm$ \textbf{3.4} \\
\specialrule{.1em}{.05em}{.05em}
\end{tabular}}\caption{Classification accuracy (\%) on social network datasets averaged over 10 runs.} 
\label{Tab:social-graph-baselines}
\end{table}

\begin{table}
\centering 
\scalebox{0.90}{\hspace{-0.35cm}\begin{tabular}{l c c c c c} 
\specialrule{.1em}{.05em}{.05em}
Method & ogbg-molhiv & ogbg-moltox21 & ogbg-moltoxcast & ogbg-ppa & ogbg-molpcba \\ [0.5ex] 
\hline 
GIN+VN \cite{hu2020open} & 75.20 $\pm$ 1.30 & 76.21 $\pm$ 0.82 & 66.18 $\pm$ 0.68 & 70.37 $\pm$ 1.07 & 27.03 $\pm$ 0.23 \\
GSN \cite{bouritsas2020improving} & 77.99 $\pm$ 1.00 & - & - & - & - \\
GraphSNN \cite{wijesinghe2021new} & 78.51 $\pm$ 1.70 & 75.45 $\pm$ 1.10 &  65.40 $\pm$ 0.71 &  70.66 $\pm$ 1.65 & 24.96 $\pm$ 1.50 \\
DS-GNN (EGO+) \cite{bevilacquaequivariant} & 77.40 $\pm$ 2.19 & 76.39 $\pm$ 1.18 & - & - & - \\
DSS-GNN (EGO+) \cite{bevilacquaequivariant} & 76.78 $\pm$ 1.66 & 77.95 $\pm$ 0.40 & - & - & - \\
POLICY-LEARN \cite{bevilacquaefficient} & 78.49 $\pm$ 1.01 & 77.36 $\pm$ 0.60 & - & - & - \\
\hline
\rowcolor{LightCyan}StructPosGSSL-SA & \textbf{78.80} $\pm$ \textbf{2.27} & \textbf{78.50} $\pm$ \textbf{2.30} & \textbf{66.50} $\pm$ \textbf{2.60} & \textbf{70.81} $\pm$ \textbf{1.45} & \textbf{27.10} $\pm$ \textbf{1.65} \\
\rowcolor{LightCyan}StructPosGSSL-FA & \textbf{79.00} $\pm$ \textbf{2.43} & \textbf{77.60} $\pm$ \textbf{2.25} & \textbf{67.00} $\pm$ \textbf{2.20} & \textbf{70.90} $\pm$ \textbf{1.86} & \textbf{27.45} $\pm$ \textbf{1.95} \\
\specialrule{.1em}{.05em}{.05em}
\end{tabular}}
\caption{Classification accuracy (\%) on OGB datasets averaged over 10 runs.
 \label{Tab:experiment_OGB}}
\end{table}

\subsection{Experiments on Synthetic Graphs}
We use 2 publicly accessible  datasets : (1) the Circular Skip Link (CSL) dataset \cite{murphy2019relational}; and (2)  SR25  \cite{abboud2020surprising}. Both benchmarks involve classifying graphs into isomorphism classes. The CSL dataset, initially presented by \cite{murphy2019relational} and frequently utilized to assess graph expressiveness \cite{dwivedi2023benchmarking}, comprises 10 isomorphism classes of 41-node 4-regular graphs, almost all of which can be distinguished by the 3-WL test. SR25 dataset \cite{abboud2020surprising} comprises 15 strongly regular graphs, each consisting of 25 nodes, which cannot be distinguished by the 3-WL test.

We compare our approach against the five baselines: GCN \cite{kipf2016semi}, GIN \cite{hou2019measuring}, 3WLGNN \cite{maron2019provably}, 3-GCN \cite{morris2019weisfeiler}, and GCN-RNI \cite{abboud2020surprising}. We use the Adam optimizer with a learning rate of 0.001, a batch size of 32, dropout of 0.7, positional encoding dimension of 6, a temperature scaling parameter $\tau$ of 0.10, and run training for 500 epochs across both datasets. We use a 3-layer MLP with a hidden dimension of 200 and a number of hops $k=3$ for both settings. We choose $\lambda_{inv}=1, \lambda_{var}=25, \lambda_{cov}=25$, and $\alpha=0.0009$ for the CSL dataset and $\lambda_{inv}=1, \lambda_{var}=24, \lambda_{cov}=24$, and $\alpha=0.005$ for the SR25 dataset. In Table \ref{Tab:experiment_expressive}, we present the average and standard deviation obtained from 10-fold cross-validation.

Note that none of the baselines achieved the best performance on both synthetic datasets we evaluated, compared to the other baselines. To address \textbf{Q3}, as shown in Table \ref{Tab:experiment_expressive}, StructPosGSSL consistently achieves the best performance on both synthetic datasets. Specifically, \emph{StructPosGSSL} improves upon the best results of the baselines by a margin of $0.8 \%$ (3WLGNN) and $93.4 \%$ (GCN, GIN, 3-GCN, and GCN-RNI) on the datasets CSL and SR25, respectively.

\subsection{Ablation Analysis of Structural and Positional Encoding}
To showcase the effectiveness of structural and positional information, we perform an ablation study on the following variants:
\begin{itemize}[leftmargin=0.6cm]
    \item POS: This variant keeps only Positional (POS) encoding.
    \item CW: This variant keeps only Closed-Walk (CW) information.
\end{itemize}

\begin{table}
\centering 
\scalebox{0.9}{\begin{tabular}{l c c} 
\specialrule{.1em}{.05em}{.05em}
Method & CSL & SR25 \\ [0.5ex] 
\hline 
StructPosGSSL-SA [POS] & 54.7 $\pm$ 3.7 & 65.3 $\pm$ 3.5 \\
StructPosGSSL-FA [POS] & 82.7 $\pm$ 2.5 & 88.3 $\pm$ 2.8 \\
\hline 
StructPosGSSL-SA [CW] & \textbf{92.0} $\pm$ \textbf{2.8} & \textbf{93.5} $\pm$ \textbf{3.7} \\
StructPosGSSL-FA [CW] & 91.3 $\pm$ 3.2 & 92.5 $\pm$ 2.8 \\
\specialrule{.1em}{.05em}{.05em}
\end{tabular}}
\caption{Ablation study. The classification accuracy (\%) for CSL/ SR25 test sets is reported. 
\label{Tab:ablation_pos_sc}}\vspace*{-0.5cm} 
\end{table}

We performed an ablation study on the \emph{StructPosGSSL} variants. The results presented in Table \ref{Tab:ablation_pos_sc} demonstrate that the closed-walk structural information plays a key role in performance. Notably, this structural information has the most significant influence, while positional information has the least impact on the CSL and SR25 datasets, as shown in Table \ref{Tab:ablation_pos_sc}.

To address the \textbf{Q4}, performance on the CSL dataset decreases by $43.9\%$ on StructPosGSSL-SA [POS], by $15.6\%$ on StructPosGSSL-FA [POS], $6.6\%$ on StructPosGSSL-SA [CW], and $7\%$ on StructPosGSSL-FA [CW], compared to the original performance shown in Table \ref{Tab:experiment_expressive}. Similarly, performance on the SR25 dataset decreases by $34.7\%$ on StructPosGSSL-SA [POS], by $11.7\%$ on StructPosGSSL-FA [POS], $6.5\%$ on StructPosGSSL-SA [CW], and $7.5\%$ on StructPosGSSL-FA [CW], compared to the original performance shown in Table \ref{Tab:experiment_expressive}.

\section{Conclusions}\label{sec:conclusions}
In conclusion, our proposed \emph{StructPosGSSL} framework effectively addresses a key limitation in Graph Self-Supervised Learning by improving the capture of topological information. Leveraging the $k$-hop message-passing mechanism of \emph{GenHopNet} and the integration of structural and positional awareness, \emph{StructPosGSSL} exceeds the expressiveness of traditional GNNs and the Weisfeiler-Lehman test. Our experimental results show that the framework delivers superior performance on graph classification tasks, enhancing accuracy while maintaining computational efficiency. This advancement significantly strengthens GSSL's capability to distinguish between graphs with similar local structures but distinct global topologies.

\textbf{Acknowledgments.} Funding for this work is from CSIRO Science Digital and Advanced Engineering Biology Future Science Platforms.

{\small
\bibliographystyle{ieee_fullname}
\bibliography{references.bib}
}

\appendix

\section{Appendices}

\subsection{Laplacian Eigenvectors for Positional Encoding}
Positional features should ideally differentiate nodes that are far apart in the graph while ensuring that nearby nodes have similar features. We use graph Laplacian eigenvectors as node positional features because they have fewer ambiguities and more accurately represent distances between nodes \cite{vaswani2017attention, dwivedi2023benchmarking}. Laplacian eigenvectors can embed graphs into Euclidean space, providing a meaningful local coordinate system while preserving the global graph structure. They are mathematically defined by the factorization of the graph Laplacian matrix as $\mathbf{L} = \mathbf{U} \Lambda \mathbf{U}^{H}$, where $\mathbf{U}=\left\{{\mathbf{u}_i}\right\}_{i=1}^{m} \in \mathbb{R}^{m}$ are orthogonal eigenvectors, $\mathbf{\Lambda}=diag\left(\left[\lambda_1,\dots,\lambda_{m}\right]\right) \in \mathbb{R}^{m\times m}$ are real eigenvalues, and $\mathbf{U}^H$ is a hermitian transpose of $\mathbf{U}$. After normalizing to unit length, eigenvectors are defined up to a factor of $\pm 1,$ leading to random sign flips during training. In our experiments, we employ the $p$ smallest non-trivial eigenvectors, with $p$ specified for each experiment. The initial positional encoding vector for each node is computed beforehand and assigned as node attributes during dataset creation.

\begin{table}[ht]
\centering 
\scalebox{0.8}{\begin{tabular}{l c c c c c c c}
\specialrule{.1em}{.05em}{.05em} 
Variants \hspace*{1.5cm}&\hspace*{0.3cm} MUTAG\hspace*{0.3cm} & \hspace*{0.3cm}PTC-MR\hspace*{0.3cm} & \hspace*{0.3cm}NCI1\hspace*{0.3cm} & \hspace*{0.3cm}PROTEINS\hspace*{0.3cm} & \hspace*{0.3cm}IMDB-B\hspace*{0.3cm} & \hspace*{0.3cm}IMDB-M\hspace*{0.3cm} \\ [0.5ex]
\hline
{NT-Xent+NoVICReg} & 88.5 $\pm$ 3.5 & 63.1 $\pm$ 3.0 & 78.3 $\pm$ 3.1 & 74.8 $\pm$ 3.5 & 73.5 $\pm$ 3.8 & 49.0 $\pm$ 3.6 \\
{NT-Xent+Inv} & 90.0 $\pm$ 3.4 & 64.3 $\pm$ 4.1 & 79.3 $\pm$ 3.2 & 71.7 $\pm$ 4.1 & 73.1 $\pm$ 4.1 & 49.8 $\pm$ 3.3 \\
{NT-Xent+Var} & 90.5 $\pm$ 3.6 & 64.8 $\pm$ 4.2 & 79.9 $\pm$ 3.6 & 72.8 $\pm$ 4.6 & 73.8 $\pm$ 4.3 & 50.3 $\pm$ 4.3 \\
{NT-Xent+Cov} & 92.0 $\pm$ 3.4 & 66.0 $\pm$ 3.2 & 81.0 $\pm$ 3.4 & 73.9 $\pm$ 4.3 & 74.2 $\pm$ 3.4 & 51.6 $\pm$ 4.1 \\
\specialrule{.1em}{.05em}{.05em}
\end{tabular}}\caption{Classification accuracy (\%) averaged over 10 runs.} 
\label{Tab:ablation}
\end{table}

\subsection{Proofs of Lemmas and Theorems}
\emph{\textbf{Theorem 1}\label{theorem-1} The following statement is true: (a) If $\sum_{k} {A}^k_{vv} \simeq_{Cycle} \sum_{k} {A}^k_{v'v'}$, then $\sum_{k} {A}^k_{vv} \simeq_{ClosedWalk} \sum_{k} {A}^k_{v'v'}$; but not vice versa.}

\begin{proof}
The implication $\sum_{k} {A}^k_{vv} \simeq_{Cycle} \sum_{k} {A}^k_{v'v'} \Rightarrow \\ \sum_{k} {A}^k_{vv} \simeq_{ClosedWalk} \sum_{k} {A}^k_{v'v'}$ is true because every cycle is a closed walk, but not every closed walk is a cycle. Thus, if two nodes are isomorphic with respect to cycles, they must also be isomorphic with respect to closed walks. The reverse implication $\sum_{k} {A}^k_{vv} \simeq_{ClosedWalk} \sum_{k} {A}^k_{v'v'} \Rightarrow \sum_{k} {A}^k_{vv} \simeq_{Cycle} \sum_{k} {A}^k_{v'v'}$ is not true because closed walks can include walks that repeat vertices or edges, which do not qualify as cycles.
\end{proof}

\emph{\textbf{Theorem 2}\label{theorem-EGMP2} Let $S$ represent a GNN with an aggregation scheme $\pi'$ delineated by Eq.~\ref{eq:1}-Eq.~\ref{eq:4}. $S$ exceeds the expressiveness of 1-WL in identifying non-isomorphic graphs, provided that $S$ operates over a sufficient number of hops, where $k > 1$, and also meets the following criteria:
\begin{itemize}
    \item[(1)] $\pi' \Big(\mathbf{h}^{(t)}_v, \{\!\!\big\{({A}_{vu}, \mathbf{h}^{(t)}_u, \mathbf{e}^{b}_{uv}, \mathbf{e}^{c}_{uv})| u \in \mathcal{N}(v) \big\}\!\!\}, \{\!\!\big\{(\tilde{{A}}^k_{vu}, \mathbf{h}^{(t)}_u )| u \in \mathcal{N}^k(v) \big\}\!\!\big\}, \{\!\!\big\{({A}^k_{vv}, \mathbf{h}^{(t)}_v\big)\big\}\!\!\}\Big)$ is injective (Eq. ~\ref{eq:gen_hop_net});
    \item[(2)] The graph-level readout function of $S$ is injective (Eq. ~\ref{eq:gen_hop_net2}).
\end{itemize}}

\begin{proof}
For the proof, we proceed in two steps. First, we assume the existence of two graphs $G_1$ and $G_2$ that are distinguishable by 1-WL but indistinguishable by $S$, and we demonstrate a contradiction. We consider the iterations of 1-WL from $1$ to $k$, where $k$ is the number of hops. If 1-WL distinguishes $G_1$ and $G_2$ using the information up to the $k$-th iteration but $S$ cannot, it implies the existence of $k$-hop local neighborhood subgraphs $\mathcal{G}_i$ and $\mathcal{G}_j$ with different multisets of $\mathbf{W}_1 \in \mathcal{W}_1$, $\mathbf{W}_2 \in \mathcal{W}_2$, $\mathbf{W}_3 \in \mathcal{W}_3$. However, by the injectiveness property of $\pi'$, $S$ should yield different tuple $(\mathbf{W}_1, \mathbf{W}_2, \mathbf{W}_3)$ for $\mathcal{G}_i$ and $\mathcal{G}_j$, contradicting the assumption. In the second step, we prove the existence of at least two graphs distinguishable by $S$ but indistinguishable by 1-WL. This step involves providing specific examples of such graphs, illustrating $S$'s enhanced expressiveness compared to 1-WL. By completing these steps, we establish the validity of Theorem \ref{theorem-EGMP2}, confirming that under the specified conditions, $S$ indeed surpasses the expressiveness of 1-WL in identifying non-isomorphic graphs.
\end{proof}


\emph{\textbf{Lemma 1}\label{lem:lem-1b}
Given two distinct pairs of multisets $\mathbf{W}_1, \mathbf{W}_1^{'} \in \mathcal{W}_1$, $\mathbf{W}_2, \mathbf{W}_2^{'} \in \mathcal{W}_2$, $\mathbf{W}_3, \mathbf{W}_3^{'} \in \mathcal{W}_3$, there exists a function $f$ such that the aggregation function $\pi(\mathbf{W}_1, \mathbf{W}_2, \mathbf{W}_3)$ and $\pi(\mathbf{W}_1^{'}, \mathbf{W}_2^{'}, \mathbf{W}_3^{'})$ defined as $\pi(\mathbf{W}_1, \mathbf{W}_2, \mathbf{W}_3)=\sum_{\mathbf{w}_1 \in \mathbf{W}_1} f(\mathbf{w}_1)+\sum_k \Big(f(\mathbf{w}_3)+\sum_{\mathbf{w}_2 \in \mathbf{W}_2} f(\mathbf{w}_2)\Big)$ and $\pi(\mathbf{W}_1^{'}, \mathbf{W}_2^{'}, \mathbf{W}_3^{'})=\sum_{\mathbf{w}_1^{'} \in \mathbf{W}_1^{'}} f(\mathbf{w}_1^{'})+\sum_k \Big(f(\mathbf{w}_3^{'})+\sum_{\mathbf{w}_2^{'} \in \mathbf{W}_2^{'}} f(\mathbf{w}_2^{'})\Big)$ are unique, respectively.}

\begin{proof}
Since $\mathcal{W}_1$, $\mathcal{W}_2$, and $\mathcal{W}_3$ are countable, there must exist three functions $\psi_1: \mathcal{W}_1\rightarrow \mathbb{N}_{odd}$ mapping $\mathbf{w}_1 \in \mathcal{W}_1$ to odd natural numbers, and $\psi_2: \mathcal{W}_2\rightarrow \mathbb{N}_{even}$ and $\psi_3: \mathcal{W}_3\rightarrow \mathbb{N}_{even}$ mapping $\mathbf{w}_2 \in \mathcal{W}_2$ and $\mathbf{w}_3 \in \mathcal{W}_3$ to even natural numbers, respectively. For any pair of multisets $(\mathbf{W}_1, \mathbf{W}_2, \mathbf{W}_3)$, given that the cardinalities of $\mathbf{W}_1, \mathbf{W}_2$, and $\mathbf{W}_3$ are bounded, there must be a natural number $N$ such that $|\mathbf{W}_1| < N$, $|\mathbf{W}_2| < N$ and $|\mathbf{W}_3| < N$. Consider a prime number $P > 3N$. Define the function $f$ such that $f(\mathbf{w}_1, \mathbf{w}_2, \mathbf{w}_3)=P^{-\psi_1(\mathbf{w}_1)} + P^{-\psi_2(\mathbf{w}_2)} + P^{-\psi_3(\mathbf{w}_3)}$. Then, the aggregation functions $\pi(\mathbf{W}_1, \mathbf{W}_2, \mathbf{W}_3)$ and $\pi(\mathbf{W}_1^{'}, \mathbf{W}_2^{'}, \mathbf{W}_3^{'})$ are unique for each distinct pair of multisets because the sum of these functions will be unique for distinct pairs of multisets by the properties of prime numbers and the unique mappings $\psi_1$, $\psi_2$ and $\psi_3$.
\end{proof}


\emph{\textbf{Lemma 2}\label{lem:lem-2b}
Expanding upon Lemma 1, we introduce an extended aggregation function $\pi'(\mathbf{h}_v \mathbf{W}_1, \mathbf{W}_2, \mathbf{W}_3)$, which incorporates the feature vector of the central node $h_v$ and the multisets $\mathbf{W}_1 \in \mathcal{W}_1$, $\mathbf{W}_2 \in \mathcal{W}_2$, and $\mathbf{W}_3 \in \mathcal{W}_3$. There exists a function $f$ such that $\pi(\mathbf{h}_v, \mathbf{W}_1, \mathbf{W}_2, \mathbf{W}_3)=(1+\epsilon) f(\mathbf{h}_v)+\sum_{\mathbf{w}_1\in \mathbf{W}_1} f(\mathbf{w}_1)+\sum_k \Big(f(\mathbf{w}_3)+\sum_{\mathbf{w}_2 \in \mathbf{W}_2} f(\mathbf{w}_2)\Big)$ is unique for any distinct quadruple $(\mathbf{h}_v, \mathbf{W}_1, \mathbf{W}_2, \mathbf{W}_3)$, where $\mathbf{h}_v\in \mathcal{H}$, $\mathbf{w}_3 \in \mathbf{W}_3$, and $\epsilon$ is an arbitrary real number.}

\begin{proof}
Let $(\mathbf{h}_v, \mathbf{W}_1, \mathbf{W}_2, \mathbf{W}_3)$ and $(\mathbf{h}_v^{'}, \mathbf{W}_1^{'}, \mathbf{W}_2^{'}, \mathbf{W}_3^{'})$ be two different tuples. Then, there are two cases:
\begin{itemize}
    \item[(1)] When $\mathbf{h}_v$ = $\mathbf{h}_v^{'}$ but $(\mathbf{h}_v, \mathbf{W}_1, \mathbf{W}_2, \mathbf{W}_3)\neq (\mathbf{h}_v^{'}, \mathbf{W}_1^{'}, \mathbf{W}_2^{'}, \mathbf{W}_3^{'})$, by Lemma~\ref{lem:lem-1b}, we know that $\sum_{\mathbf{w}_1\in \mathbf{W}_1} f(\mathbf{w}_1)+\\ \sum_k \Big(f(\mathbf{w}_3)+\sum_{\mathbf{w}_2 \in \mathbf{W}_2} f(\mathbf{w}_2)\Big) \neq \sum_{\mathbf{w}_1^{'} \in \mathbf{W}_1^{'}} f(\mathbf{w}_1^{'})+\sum_k \Big(f(\mathbf{w}_3^{'})+\sum_{\mathbf{w}_2^{'} \in \mathbf{W}_2^{'}} f(\mathbf{w}_2^{'})\Big)$. Thus, $\pi'(\mathbf{h}_v, \mathbf{W}_1, \mathbf{W}_2, \mathbf{W}_3) \neq \pi'(\mathbf{h}_v^{'}, \mathbf{W}_1^{'}, \mathbf{W}_2^{'}, \mathbf{W}_3^{'})$. 
    \item[(2)] When $\mathbf{h}_v \neq \mathbf{h}_v^{'}$, we prove $\pi'(\mathbf{h}_v, \mathbf{W}_1, \mathbf{W}_2, \mathbf{W}_3) \neq \\ \pi'(\mathbf{h}_v^{'}, \mathbf{W}_1^{'}, \mathbf{W}_2^{'}, \mathbf{W}_3^{'})$ by contradiction. Assume that \\ $\pi'(\mathbf{h}_v, \mathbf{W}_1, \mathbf{W}_2, \mathbf{W}_3)= \pi'(\mathbf{h}_v^{'}, \mathbf{W}_1^{'}, \mathbf{W}_2^{'}, \mathbf{W}_3^{'})$. Then, we have:
    \begin{equation*}
        \begin{split}
        (1+\epsilon) f(\mathbf{h}_v) + \sum_{\mathbf{w}_1 \in \mathbf{W}_1} f(\mathbf{w}_1) + \sum_k \Big(f(\mathbf{w}_3) + \sum_{\mathbf{w}_2 \in \mathbf{W}_2} f(\mathbf{w}_2)\Big) &= \\
        (1+\epsilon) f(\mathbf{h}_v^{'}) + \sum_{\mathbf{w}_1^{'} \in \mathbf{W}_1^{'}} f(\mathbf{w}_1^{'}) + \\ \sum_k \Big(f(\mathbf{w}_3^{'}) + \sum_{\mathbf{w}_2^{'} \in \mathbf{W}_2^{'}} f(\mathbf{w}_2^{'})\Big).
        \end{split}
    \end{equation*}
    This gives us the following equation:
    \begin{equation*}
        \begin{split}
            (1+\epsilon)\Big(f(\mathbf{h}_v)-f(\mathbf{h}_v^{'})\Big) &=
            \sum_{\mathbf{w}_1 \in \mathbf{W}_1} f(\mathbf{w}_1)-\sum_{\mathbf{w}_1^{'} \in \mathbf{W}_1^{'}} f(\mathbf{w}_1^{'}) \\ &+
            \Big(\sum_k \Big(f(\mathbf{w}_3)+\sum_{\mathbf{w}_2 \in \mathbf{W}_2} f(\mathbf{w}_2)\Big) \\ &- \Big(\sum_k \Big(f(\mathbf{w}_3^{'})+\sum_{\mathbf{w}_2^{'} \in \mathbf{W}_2^{'}} f(\mathbf{w}_2^{'})\Big).
        \end{split}
    \end{equation*}
    If $\epsilon$ is an irrational number, the left-hand side of the equation is irrational, while the right-hand side is rational, leading to a contradiction. Therefore, $\pi'(\mathbf{h}_v, \mathbf{W}_1, \mathbf{W}_2, \mathbf{W}_3) \neq \pi'(\mathbf{h}_v^{'}, \mathbf{W}_1^{'}, \mathbf{W}_2^{'}, \mathbf{W}_3^{'})$.
\end{itemize}
\end{proof}

\emph{\textbf{Corollary 1}
GenHopNet exhibits greater expressiveness compared to 1-WL when evaluating non-isomorphic graphs.}
\begin{proof}
We demonstrate this theorem by proving that \emph{GenHopNet} is a GNN that meets the conditions specified in Theorem 2. For the first condition, consider the two graphs depicted in Figure \ref{fig:model-architecture}(b). \emph{GenHopNet} can differentiate these graphs as $\{\!\!\big\{(A_{vu},\mathbf{h}^{(t)}_u,\mathbf{e}^{b}_{uv}, \mathbf{e}^{c}_{uv})| u \in \mathcal{N}(v) \big\}\!\!\} \neq \{\!\!\big\{(A_{v'u'},\mathbf{h}^{(t)}_{u'},\mathbf{e}^{b}_{u'v'}, \mathbf{e}^{c}_{u'v'})| u' \in \mathcal{N}(v') \big\}\!\!\}, \{\!\!\big\{(\tilde{A}^k_{vu},\mathbf{h}^{(t)}_u )| u \in \mathcal{N}^k(v) \big\}\!\!\big\} \neq \{\!\!\big\{(\tilde{A}^k_{v'u'},\mathbf{h}^{(t)}_{u'})| u' \in \mathcal{N}^k(v') \big\}\!\!\big\}$, and $\{\!\!\big\{({A}^k_{vv}, \mathbf{h}^{(t)}_v\big)\big\}\!\!\} \neq \{\!\!\big\{({A}^k_{v'v'}, \mathbf{h}^{(t)}_{v'}\big)\big\}\!\!\}$. For the second condition, leveraging Lemmas 1 and 2, along with the fact that an MLP can serve as a universal approximator \cite{xu2018powerful} to model and learn the function $f$, we establish that \emph{GenHopNet} also satisfies this condition.
\end{proof}

\emph{\textbf{Theorem 3}\label{theorem-4} The {StructPosGSSL}
 is more expressive than subgraph MPNNs in distinguishing certain non-isomorphic graphs.}

\begin{figure}
    \centering
    \includegraphics[width=0.5\textwidth]{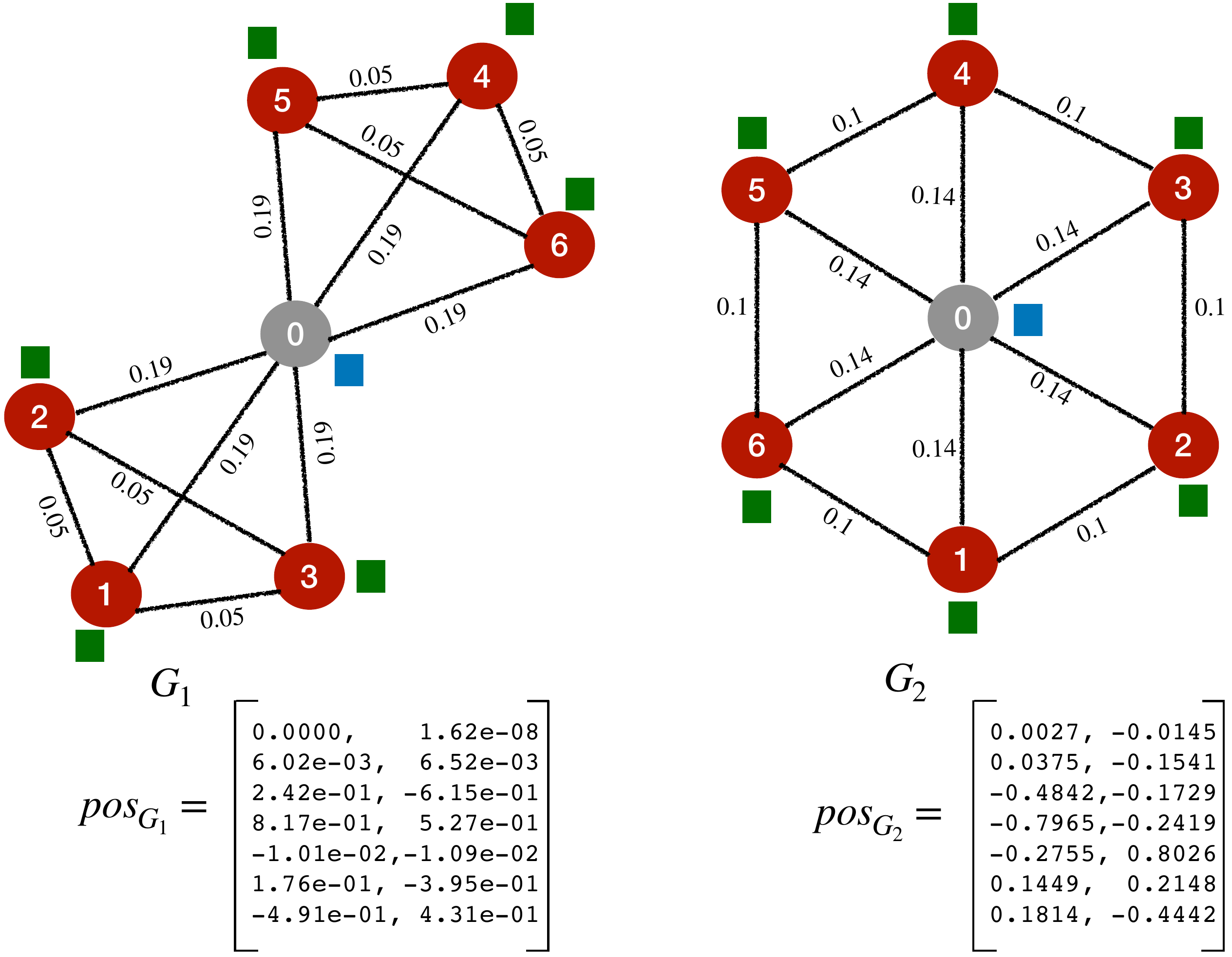}
    \caption{A pair of non-isomorphic graphs where the colored square box on each node represents the feature representations derived from closed-walk information. The two middle nodes (colored gray) in graphs $G_1$ and $G_2$ cannot be distinguished using only closed-walk information (up to $k=3$), as they receive the same representation (colored blue). However, by incorporating positional information along with EB attributes, we can successfully distinguish these nodes.}
    \label{fig:non_isomorphic_graphs}
\end{figure}

\begin{proof}
To prove Theorem 3, we consider the two non-isomorphic graphs $G_1$ and $G_2$ shown in Figure ~\ref{fig:non_isomorphic_graphs}. Let $v \in G_1$ and $v' \in G_2$ be the middle nodes in each graph. In the case of Subgraph MPNNs (e.g., \cite{you2021identity, cotta2021reconstruction, zhang2021nested}), the aggregation function over the neighborhoods of $v$ and $v'$ fails to differentiate between the two nodes. This is because Subgraph MPNNs rely on local subgraphs, and the structural features and neighborhood-based information are symmetric for $v$ and $v'$.

Let us first consider the structural encoder (\ie, \emph{GenHopNet}) with only closed-walk information up to $k=3$, without EB attributes $e^{c}_{uv}$. In this case, the node representations for $v$ and $v'$ generated by the structural encoder using closed-walks are identical, since $\\ \{\!\!\big\{(A_{vu},\mathbf{h}^{(t)}_u,\mathbf{e}^{b}_{uv})| u \in \mathcal{N}(v) \big\}\!\!\} = \{\!\!\big\{(A_{v'u'},\mathbf{h}^{(t)}_{u'},\mathbf{e}^{b}_{u'v'})| u' \in \mathcal{N}(v') \big\}\!\!\}, \\ \{\!\!\big\{(\tilde{A}^k_{vu},\mathbf{h}^{(t)}_u )| u \in \mathcal{N}^k(v) \big\}\!\!\big\} = \{\!\!\big\{(\tilde{A}^k_{v'u'},\mathbf{h}^{(t)}_{u'})| u' \in \mathcal{N}^k(v') \big\}\!\!\big\}$, and $\{\!\!\big\{({A}^k_{vv}, \mathbf{h}^{(t)}_v\big)\big\}\!\!\} = \{\!\!\big\{({A}^k_{v'v'}, \mathbf{h}^{(t)}_{v'}\big)\big\}\!\!\}$. This indicates that the closed-walk information alone is insufficient to distinguish these non-isomorphic graphs. Now, we show how \emph{StructPosGSSL}, when enhanced with positional encodings and $\mathbf{e}^{c}_{uv}$, can differentiate between the non-isomorphic graphs $G_1$ and $G_2$. Let $\mathbf{h}^{(t)}_{u, pos}$ be the positional encoding of node $u$. When positional encodings are combined with $\mathbf{e}^{c}_{uv}$, the aggregation function can distinguish these non-isomorphic graph pairs. Using Lemmas 1 and 2, we know that the aggregation function is still injective when positional encodings and $\mathbf{e}^{c}_{uv}$ are included. Thus, for the middle nodes of each graph, we have $\{\!\!\big\{(A_{vu},\mathbf{h}^{(t)}_{u, pos}, \mathbf{e}^{b}_{uv}, \mathbf{e}^{c}_{uv})| u \in \mathcal{N}(v) \big\}\!\!\} \neq \{\!\!\big\{(A_{v'u'},\mathbf{h}^{(t)}_{u', pos},\mathbf{e}^{b}_{u'v'}, \mathbf{e}^{c}_{u'v'})| u' \in \mathcal{N}(v') \big\}\!\!\}$. Therefore, the \emph{StructPosGSSL} with positional encodings and EB attributes yields different representations for $v$ and $v'$, even though they were previously indistinguishable.
\end{proof}

\subsection{Ablation Analysis of Loss Function}
To showcase the effectiveness of each element in the loss function, we perform an ablation study on the following variants:
\begin{itemize}
    \item NoVICReg: This variant excludes the VICReg regularization term from the overall loss.
    \item Inv: This variant keeps only the Invariance term in the VICReg regularization term.
    \item Var: This variant keeps only the Variance term in the VICReg regularization term.
    \item Cov: This variant keeps only the Covariance term in the VICReg regularization term.
\end{itemize}

We conducted the ablation study on the \emph{StructPosGSSL-SA} variant. The results shown in Table \ref{Tab:ablation} indicate that the Invariance, Variance, and Covariance terms are crucial to the performance. Specifically, the covariance term has the greatest impact on performance, whereas the invariance term has the least effect across all datasets, as detailed in Table \ref{Tab:ablation}. Specifically, as shown in Table \ref{Tab:ablation}, performance on graphs decreases by $3.5\%$ to $5.1\%$ with the NT-Xent+NoVICReg loss function, by $2.4\%$ to $4.1\%$ with NT-Xent+Inv, by $1.7\%$ to $3.0\%$ with NT-Xent+Var, and by $1.0\%$ to $1.9\%$ with NT-Xent+Cov, compared to the combined NT-Xent+VICReg loss function. 

\subsection{Comparison under Different $\alpha$}
\begin{figure}
    \centering
    \includegraphics[width=0.5\textwidth]{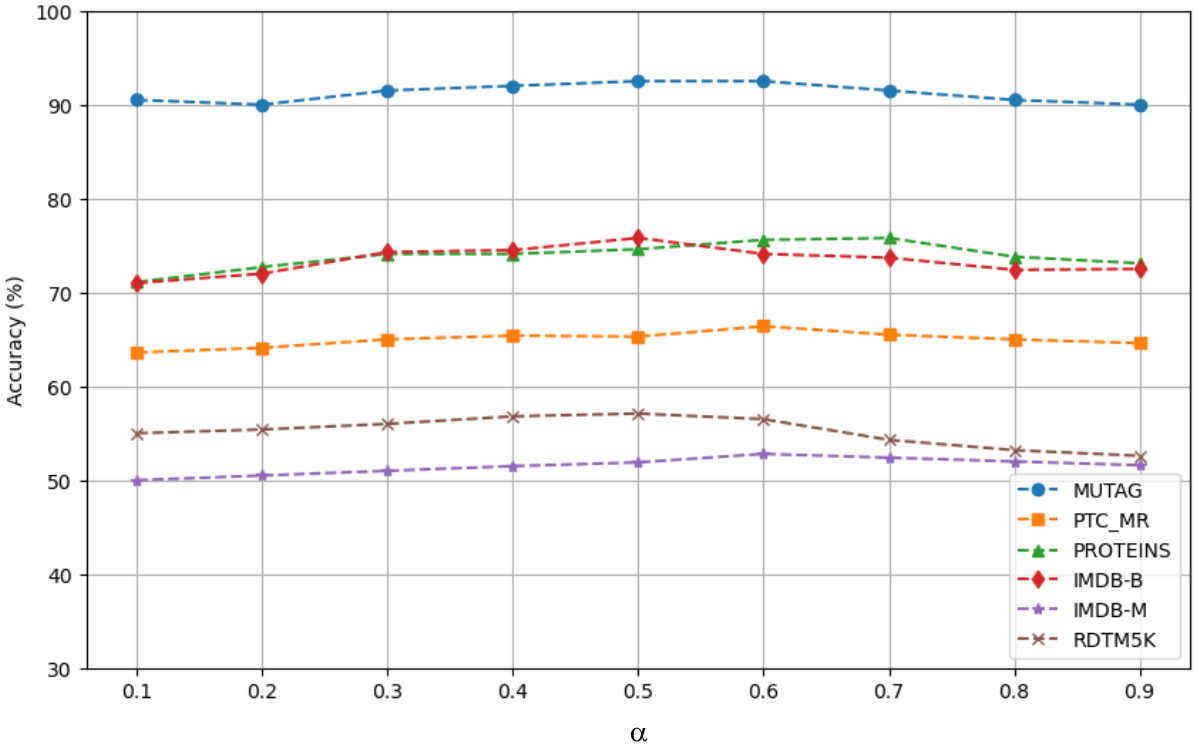}
    \vspace{-0.1cm}
    \caption{Accuracy (\%) of StructPosGSSL-SA under different $\alpha$ values.}
    \label{fig:comparison-mu}
\end{figure}

To evaluate the impact of the regularization term  $\alpha$ on the performance of our \emph{StructPosGSSL} framework, we conduct experiments by evaluating \emph{StructPosGSSL-SA} across six datasets (MUTAG, PTC-MR, PROTEINS, IMDB-B, IMDB-M, and RDT5K), using varying values for the regularization term $\alpha=\{0.1, 0.2, \ldots, 0.9\}$. For this experimental setup, we use the same hyperparameter configuration for each dataset as described in Section \ref{sec:small-graphs}. Figure \ref{fig:comparison-mu} represents the experimental results. In our experiments, we observed that setting $\alpha$ either too low or too high leads to suboptimal performance. To achieve better results, it is essential to select an intermediate value for $\alpha$, as this provides a balance that optimizes the \emph{StructPosGSSL}'s performance.

\subsection{Ablation Analysis of GenHopNet with traditional SSL}
To demonstrate the efficacy of \emph{GenHopNet} alongside traditional SSL, we conduct an ablation study on the following variations:
\begin{itemize}
    \item \emph{GenHopNet}+NT-Xent: Combines our \emph{GenHopNet} with NT-Xent loss.
    \item GCN+NT-Xent: Replaces \emph{GenHopNet} with GCN while retaining NT-Xent loss.
\end{itemize}

\begin{table}[ht]
    \centering 
    \scalebox{0.9}{\begin{tabular}{l c c c c c c c}
    \specialrule{.1em}{.05em}{.05em} 
    Variants \hspace*{1.5cm}&\hspace*{0.3cm} ogbg-molhiv\hspace*{0.3cm} & \hspace*{0.3cm}ogbg-motox21\hspace*{0.3cm} & \hspace*{0.3cm}ogbg-moltoxcast\hspace*{0.3cm} \\ [0.5ex]
    \hline
    {\emph{GenHopNet}+NT-Xent}-SA & \textbf{74.0} $\pm$ \textbf{2.7} & \textbf{73.5} $\pm$ \textbf{3.1} & \textbf{62.3} $\pm$ \textbf{3.2} \\
    {\emph{GenHopNet}+NT-Xent}-FA & \textbf{73.3} $\pm$ \textbf{3.2} & \textbf{72.8} $\pm$ \textbf{3.3} & \textbf{62.9} $\pm$ \textbf{3.5} \\
    {GCN+NT-Xent}-SA & 69.5 $\pm$ 2.3 & 69.1 $\pm$ 3.2 & 58.3 $\pm$ 3.6 \\
    {GCN+NT-Xent}-FA & 70.9 $\pm$ 2.6 & 69.6 $\pm$ 3.3 & 57.5 $\pm$ 3.1 \\
    \specialrule{.1em}{.05em}{.05em}
    \end{tabular}}\caption{Classification accuracy (\%) averaged over 10 runs.} 
    \label{Tab:ablation_gen_trad}
\end{table}
The results in Table ~\ref{Tab:ablation_gen_trad} demonstrate that GenHopNet+NT-Xent consistently outperforms GCN+NT-Xent across various datasets. For instance, on ogbg-molhiv, \emph{GenHopNet}+NT-Xent achieves a classification accuracy of 74.0\%, compared to 69.5\% with GCN+NT-Xent. Similarly, on ogbg-moltox21 and ogbg-moltoxcast, the improvements are evident with accuracies of 73.5\% vs 69.1\% and 62.3\% vs 58.3\%, respectively. These results highlight the superior expressiveness of \emph{GenHopNet}, which captures complex structural patterns through $k$-hop message passing and edge-based centrality measures, enabling more discriminative and robust graph representations in self-supervised learning setups.

\subsection{Comparison of NT-Xent and VICReg Loss Terms}
To showcase the complementary nature of NT-Xent and VICReg loss terms, we perform an ablation study on the following variants:
\begin{itemize}
    \item No VICReg: excludes VICReg term from total loss.
    \item No NT-Xent: excludes NT-Xent term from total loss.
\end{itemize}

\begin{table}[ht]
\centering 
\scalebox{0.9}{\begin{tabular}{l c c c c}
\specialrule{.1em}{.05em}{.05em} 
Variants \hspace*{1.5cm}&\hspace*{0.3cm}ogbg-molhiv\hspace*{0.3cm} & \hspace*{0.3cm}IMDB-B\hspace*{0.3cm} & \hspace*{0.3cm}MUTAG\hspace*{0.3cm} & \hspace*{0.3cm}PTC-MR\hspace*{0.3cm} \\ [0.5ex]
\hline
{NT-Xent+NoVICReg}-SA & \textbf{75.01} $\pm$ \textbf{2.51} & \textbf{72.34} $\pm$ \textbf{3.30} & 86.23 $\pm$ 3.61 & 61.36 $\pm$ 3.50 \\
{NT-Xent+NoVICReg}-FA & \textbf{74.62} $\pm$ \textbf{2.83} & \textbf{71.92} $\pm$ \textbf{3.13} & 86.01 $\pm$ 
3.32 & 61.43 $\pm$ 3.37 \\
\hline
{VICReg+NoNT-Xent}-SA & 71.65 $\pm$ 2.65 & 69.36 $\pm$ 3.51 & \textbf{89.33} $\pm$ \textbf{3.50} & \textbf{64.21} $\pm$ \textbf{3.00} \\
{VICReg+NoNT-Xent}-FA & 70.31 $\pm$ 2.13 & 68.51 $\pm$ 3.02 & \textbf{89.10} $\pm$ \textbf{3.42} & \textbf{64.32} $\pm$ \textbf{3.51} \\
\hline
{NT-Xent+VICReg}-SA & \textbf{78.80} $\pm$ \textbf{2.27} & \textbf{75.50} $\pm$ \textbf{3.30} & \textbf{93.00} $\pm$ \textbf{5.30} & \textbf{67.10} $\pm$ \textbf{4.90} \\
{NT-Xent+NoVICReg}-FA & \textbf{79.00} $\pm$ \textbf{2.43} & \textbf{75.10} $\pm$ \textbf{3.10} & \textbf{91.50} $\pm$ \textbf{2.50} & \textbf{66.50} $\pm$ \textbf{3.70} \\
\specialrule{.1em}{.05em}{.05em}
\end{tabular}}\caption{Classification accuracy (\%) averaged over 10 runs.} 
\label{Tab:ablation_nt_vic}
\end{table}

\begin{table}[ht]
    \centering 
    \scalebox{0.65}{\begin{tabular}{l c c c c c c c}
    \specialrule{.1em}{.05em}{.05em} 
    Variants \hspace*{1.5cm}&\hspace*{0.3cm} MUTAG\hspace*{0.3cm} & \hspace*{0.3cm}PTC-MR\hspace*{0.3cm} & \hspace*{0.3cm}NCI1\hspace*{0.3cm} & \hspace*{0.3cm}PROTEINS\hspace*{0.3cm} & \hspace*{0.3cm}IMDB-B\hspace*{0.3cm} & \hspace*{0.3cm}IMDB-M\hspace*{0.3cm} \\ [0.5ex]
    \hline
    {GenHopNet+Org}-SA & \textbf{93.0} $\pm$ \textbf{5.3} & \textbf{67.1} $\pm$ \textbf{4.9} & \textbf{82.9} $\pm$ \textbf{3.8} & \textbf{75.8} $\pm$ \textbf{2.6} & \textbf{75.5} $\pm$ \textbf{3.3} & \textbf{52.8} $\pm$ \textbf{3.5} \\
    {GenHopNet+Org}-FA & 91.5 $\pm$ 2.5 & 66.5 $\pm$ 3.7 & 82.5 $\pm$ 3.5 & 75.3 $\pm$ 2.9 & 75.1 $\pm$ 3.1 & 52.3 $\pm$ 3.6 \\
    \hline
    {GenHopNet+$\mathbf{m}^{(t)}_{local\_pat}(v)$}-SA & 90.5 $\pm$ 4.2 & 64.7 $\pm$ 4.6 & 80.6 $\pm$ 3.5 & 73.5 $\pm$ 2.8 & 73.3 $\pm$ 3.5 & 50.6 $\pm$ 3.1 \\
    {GenHopNet+$\mathbf{m}^{(t)}_{local\_pat}(v)$}-FA & 89.6 $\pm$ 4.1 & 64.0 $\pm$ 4.7 & 80.4 $\pm$ 3.0 & 73.1 $\pm$ 2.5 & 73.0 $\pm$ 3.1 & 50.5 $\pm$ 3.3 \\
    \hline
    {GenHopNet+$\mathbf{m}^{(t)}_{closed\_walks}(v)+\mathbf{m}^{(t)}_{high\_ord}(v)$}-SA & 88.8 $\pm$ 4.6 & 63.6 $\pm$ 4.1 & 78.6 $\pm$ 3.6 & 71.6 $\pm$ 2.9 & 71.3 $\pm$ 3.5 & 48.5 $\pm$ 3.6 \\
    {GenHopNet+$\mathbf{m}^{(t)}_{closed\_walks}(v)+\mathbf{m}^{(t)}_{high\_ord}(v)$}-FA & 88.2 $\pm$ 4.0 & 62.9 $\pm$ 4.5 & 78.3 $\pm$ 3.0 & 71.3 $\pm$ 2.5 & 71.0 $\pm$ 3.5 & 48.4 $\pm$ 3.4 \\
    \specialrule{.1em}{.05em}{.05em}
    \end{tabular}}\caption{Classification accuracy (\%) averaged over 10 runs.} 
    \label{Tab:ablation_edge_mpnn}
\end{table}

Based on the results in Table ~\ref{Tab:ablation_nt_vic}, the NT-Xent loss performs exceptionally well on dense graphs and motif-rich datasets (\eg, IMDB-B, OGB) by effectively capturing global structures and topological motifs. By leveraging structural encodings and positional encodings, it ensures robust graph-level embeddings that integrate both local and global features. On the other hand, the VICReg loss demonstrates its strength on sparse graphs (\eg., MUTAG, PTC-MR) by utilizing positional encodings to break graph symmetry and refining embeddings to capture subtle connectivity patterns and node-specific properties.

\subsection{Ablation Analysis of $\mathbf{m}^{(t)}_{local\_pat}$}
To evaluate the impact of the 1-hop edge-level aggregated message $\mathbf{m}^{(t)}_{local\_pat}(v)$ on the performance of our \emph{StructPosGSSL} framework, we conduct experiments by evaluating \emph{StructPosGSSL-SA} across six datasets (MUTAG, PTC-MR, NCI1, PROTEINS, IMDB-B, and IMDB-M), using the following variations:
\begin{itemize}
    \item \emph{GenHopNet}+Org: Original \emph{GenHopNet} without any modifications to message passing.
    \item \emph{GenHopNet}+$\mathbf{m}^{(t)}_{local\_pat}(v)$: \emph{GenHopNet} with only 1-hop edge-level aggregated message $\mathbf{m}^{(t)}_{local\_pat}(v)$ during message passing.
    \item \emph{GenHopNet}+$\mathbf{m}^{(t)}_{closed\_walks}(v)+\mathbf{m}^{(t)}_{high\_ord}(v)$: \emph{GenHopNet} without 1-hop edge-level aggregated message $\mathbf{m}^{(t)}_{local\_pat}(v)$, but includes the node-level $k$-hop ($k \ge 2$) aggregated messages of $\mathbf{m}^{(t)}_{closed\_walks}(v)$ and $\mathbf{m}^{(t)}_{high\_ord}(v)$.
\end{itemize}


The results in Table ~\ref{Tab:ablation_edge_mpnn} demonstrate that variant \emph{GenHopNet}+ \\ $\mathbf{m}^{(t)}_{local\_pat}(v)$ consistently outperforms variant \emph{GenHopNet}+ \\ $\mathbf{m}^{(t)}_{closed\_walks}(v)+\mathbf{m}^{(t)}_{high\_ord}(v)$ across all datasets. Therefore, the term $\mathbf{m}^{(t)}_{local\_pat}(v)$ in Eq. \ref{eq:gen_hop_net} is necessary for improving the distinguishing power of the proposed model.

\subsection{Comparison of Structural and Positional Encoders}
The apparent effectiveness of the Structural Encoder compared to the Positional Encoder depends on the characteristics of the datasets. Positional encodings excel in dense graphs (\eg, IMDB-B) by providing global positional context and breaking symmetry, thus distinguishing graphs when local structural cues alone are insufficient. In contrast, structural encodings prove more impactful in sparse graphs or domains such as molecular datasets (\eg, OGB, ZINC), where local patterns, motifs, or functional groups are key. Taken together, these two modules target complementary aspects of graph structure; Positional Encoder focuses on global relationships and symmetry-breaking, while SE emphasizes local substructures-enabling a more comprehensive approach to graph representation learning. 

In order to highlight how SE and PE encoders complement one another, we conduct an ablation study using the following variants:
\begin{itemize}[leftmargin=0.6cm]
    \item POS: This variant keeps only positional (POS) encoding.
    \item STRUCT: This variant keeps only structural (STRUCT) encoding.
\end{itemize}

\begin{table}[ht]
\centering
\scalebox{0.9}{\begin{tabular}{l c c c}
\specialrule{.1em}{.05em}{.05em}
Method & ogbg-molhiv & ogbg-moltox21 & ogbg-moltoxcast \\ [0.5ex]
\hline 
StructPosGSSL-SA [POS] & 71.01 $\pm$ 2.20 & 69.31 $\pm$ 3.10 & 58.21 $\pm$ 2.80 \\
StructPosGSSL-FA [POS] & 74.66 $\pm$ 2.10 & 71.65 $\pm$ 2.48 & 60.06 $\pm$ 2.65 \\
\hline 
StructPosGSSL-SA [STRUCT] & \textbf{77.32} $\pm$ \textbf{2.65} & \textbf{75.40} $\pm$ \textbf{3.15} & \textbf{63.15} $\pm$ \textbf{2.70} \\
StructPosGSSL-FA [STRUCT] & 76.90 $\pm$ 2.30 & 74.93 $\pm$ 2.85 & 62.66 $\pm$ 2.37 \\
\specialrule{.1em}{.05em}{.05em}
\end{tabular}}
\caption{Ablation study classification accuracy (\%) for OGB datasets.
\label{Tab:ablation_pos_sc_1}}
\end{table}

The results in the Table ~\ref{Tab:ablation_pos_sc_1} for molecular graphs demonstrate that closed-walk structural information, captured by the structural encoder, is more crucial than the positional encoder when substructure patterns such as cycles and motifs are important, as seen in molecular graphs like ogbg-molhiv, ogbg-motox21, and ogbg-moltoxcast. Structural encoder excels at capturing closed-walk counts, which are essential for distinguishing structural patterns, especially cycles and motifs that are central to the expressiveness of molecular graph representations. While positional encoder offer valuable global context and assist in differentiating symmetric nodes, their impact is less pronounced in molecular classification tasks, where the dominance of structural patterns outweighs the need for positional information.

\begin{table}[ht]
\centering
\scalebox{0.9}{\begin{tabular}{l c c}
\specialrule{.1em}{.05em}{.05em}
Method & IMDB-B \\ [0.5ex]
\hline 
StructPosGSSL-SA [POS] & \textbf{72.85} $\pm$ \textbf{3.30} \\
StructPosGSSL-FA [POS] & 72.05 $\pm$ 3.08 \\
\hline 
StructPosGSSL-SA [STRUCT] & 69.50 $\pm$ 3.25 \\
StructPosGSSL-FA [STRUCT] & 69.03 $\pm$ 2.97 \\
\specialrule{.1em}{.05em}{.05em}
\end{tabular}}
\caption{Ablation study classification accuracy (\%) for social network datasets.
\label{Tab:ablation_pos_sc_2}}
\end{table}

The results in the Table ~\ref{Tab:ablation_pos_sc_2} for dense graphs highlight the more significant impact of positional encoder on model performance compared to structural encoder. This is because PE captures global positional relationships and aids in breaking symmetries, providing an advantage in graphs where local structural patterns are not enough. In particular, in datasets like IMDB-B, characterized by dense connectivity, PE ensures that nodes with similar local neighborhoods but different global roles can be effectively distinguished, thus enhancing the model's ability to differentiate between nodes in complex graph structures.

\begin{table}[ht]
\centering
\scalebox{0.90}{\begin{tabular}{l c}
\specialrule{.1em}{.05em}{.05em}
Method & ZINC12K (MAE) \\ [0.5ex]
\hline 
PathNN & 0.090 $\pm$ 0.004 \\
G3N & 0.128 $\pm$ 0.015 \\ 
CIN & 0.079 $\pm$ 0.006 \\ 
$I^2$-GNN & 0.083 $\pm$ 0.001 \\
Moment-GNN & 0.110 $\pm$ 0.005 \\
K-Subgraph SAT & 0.094 $\pm$ 0.008 \\
GRIT & 0.059 $\pm$ 0.002 \\
GPS & 0.070 $\pm$ 0.004 \\
Specformer & 0.066 $\pm$ 0.003 \\
\hline
StructPosGSSL-SA & \textbf{0.050} $\pm$ \textbf{0.009} \\
StructPosGSSL-FA & \textbf{0.058} $\pm$ \textbf{0.007} \\
\specialrule{.1em}{.05em}{.05em}
\end{tabular}}
\caption{Classification accuracy (\%) on the ZINC12K dataset.
\label{Tab:experiment_expressive_zinc}}
\end{table}

\subsection{Experiments on ZINC Dataset}
The ZINC dataset \cite{wang2023mathscr} is composed of 250K molecules, from which 12K are chosen for the solubility regression task under certain constraints. This dataset allows for a comprehensive assessment of the model's ability to capture both local and global graph properties. Our experimental design adheres to the methodology outlined in \cite{dwivedi2023benchmarking}.

We compare our method against eight baseline approaches: PathNN \cite{michel2023path}, G3N \cite{wang2023mathscr}, CIN \cite{bodnar2021cellular}, $I^2$-GNN \cite{huangboosting}, Moment-GNN \cite{kanatsouliscounting}, K-Subgraph SAT \cite{chen2022structure}, GRIT \cite{ma2023graph}, GPS \cite{rampavsek2022recipe}, and Specformer \cite{bospecformer}.

For the ZINC12K dataset in Table ~\ref{Tab:experiment_expressive_zinc}, our method achieves strong performance compared to other baselines. This success can be attributed to the ability of our structural encoding to effectively capture rich local topological features, such as cycles and motifs, which are highly correlated with molecular classes and attributes. Additionally, the integration of global positional relationships further enhances the model's expressiveness, enabling it to better distinguish subtle molecular graph patterns and achieve better results.

\end{document}